\documentclass[12pt]{article}

\usepackage[latin1]{inputenc}
\usepackage{amsfonts}
\usepackage{latexsym,amssymb}
\usepackage{amsmath,amsthm}
\usepackage{enumerate}
\usepackage{stmaryrd}  
\usepackage[colors]{optsys}

\newcommand{\barbar}[1]{\bar{\bar{#1}}}

\newcommand{\proba}{p}
\newcommand{\probabis}{q}
\newcommand{\Critical}{\proba_c}
\newcommand{\good}{\mathtt{G}}
\newcommand{\bad}{\mathtt{B}}

\newcommand{\information}{{\cal O}}
\newcommand{\discount}{\rho}
\newcommand{\utility}{U}
\newcommand{\util}{{\mathcal U}}

\newcommand{\Try}{\varepsilon} 
\newcommand{\Avoid}{\alpha} 

\newcommand{\SIMPLEX}{\Sigma}
\newcommand{\Bernoulli}[2]{\mathcal{B}\np{#1,#2}}   
\newcommand{\prior}{\InformationState}

\renewcommand{\control}{v}
\renewcommand{\Control}{V}

\renewcommand{\HISTORY}{\mathbb{H}_{\infty}}
\newcommand{\InformationState}{\pi}
\newcommand{\INFORMATIONSTATE}{\Pi}
\newcommand{\kernel}[3]{k_{#1}\np{#2 \mid #3}}
\newcommand{\observation}{y}
\newcommand{\Observation}{Y}
\newcommand{\OBSERVATION}{\mathbb{Y}}
\renewcommand{\strategy}{{\cal S}}

\renewcommand{\bar}{\overline}
\newcommand{\CITE}[1]{\cite{#1}}

\title{Rationally Biased Learning}

\author{Michel \textsc{De Lara}\\
  CERMICS, Ecole des Ponts, Marne-la-Vall\'ee, France\\
E-mail: michel.delara@enpc.fr}

\begin{document}

\maketitle

\begin{abstract}
Humans display a tendency to pay more attention to bad outcomes,
often in a disproportionate way relative to their statistical occurrence.
They also display euphorism, as well as 
a preference for the current state of affairs (status quo bias).
Based on the analysis of optimal solutions of infinite horizon
stationary optimization problems under imperfect state observation,
we show that such human perception and decision biases
can be grounded in a form of rationality (optimality).
We also provide conditions (boundaries) for their possible occurence
and an analysis of their robustness.
Thus, biases can be the product of rational behavior.
\end{abstract}


\section{Introduction}
\label{Introduction}

When we perceive sounds, we overestimate the change in level of rising level 
tones relative to equivalent falling level tones \CITE{Neuhoff:1998}.
When we assess pros and cons in decision making, we weigh losses more than 
gains \CITE{Kahneman-Tversky:1979}.
We are more frightened by a snake or a spider than by a passing car or an
electrical shuffle.
Such human assessments are qualified of biases, because they depart 
from physical measurements or objective statistical estimates.
Thus, there is ``bias'' when a behavior is not aligned 
with a given ``rationality benchmark'' (like expected utility theory),
as documented in the ``heuristics and biases'' literature
\CITE{Kahneman-Slovic-Tversky:1982,Gilovich-Griffin-Kahneman:2002}.

However, if such biases are found consistently in human behavior, 
they must certainly have a reason. Some scholars (see
\CITE{Gigerenzer:2004,Gigerenzer:2008a,Hutchinson-Gigerenzer:2005}) claim that
those ``so-called bias'' were in fact advantageous in the type of environment
where our ancestors lived and thrived 
(ecological or, rather, evolutionary, validity 
\CITE{Barkow-Cosmides-Tooby:1992,Boutang-DeLara:2015,Boutang-DeLara:2019}).
In this conception, the benchmark should be a measure of fitness
reflecting survival and reproduction abilities, and the ``bias'' can be 
explained in two ways.
\begin{itemize}
\item 
\emph{Bias by mismatch}.
The bias can result from a timelag: because the modern environment has departed so
much from the environment in which natural selection had time to shape our minds,
human behavior displays a mismatch (cars are objectively more dangerous
than spiders in our modern environment).
\item 
\emph{Bias by design}.
But the bias can be the feature of an optimal strategy where optimality is
measured in fitness (the genes of those who accurately estimated the change in level of rising level 
tones have, more often than the ``overestimaters'', finished in the stomach of a predator). 
\end{itemize}
This last conception of ``bias by design'' is reflected, for example, in
\CITE{Haselton-Nettle:2006}.
In their attempt to understand ``how natural selection engineers psychological adaptations for
judgment under uncertainty'', Haselton and Nettle consider an individual who has
to decide between a safe option (one known payoff) and a risky option (known
bad and good payoffs).
They define a critical probability and observe that, if the bad outcome has 
higher probability, the (optimal) individual
should avoid taking risks and select the safe option.
The interesting point is that the critical probability is the ratio of
the difference between good and safe payoffs over
the difference between good and bad payoffs. 
As a consequence, the higher the latter difference, the more
the individual should avoid taking risks.
%
%
The general conclusion is nicely expressed by Martie G. Haselton 
(on her personal webpage) when she claims that 
``selection has led to adaptations that are biased by design and 
functioned to help ancestral humans avoid particularly costly errors''
and that 
``when the costs of false positive and false negative errors were asymmetrical over evolutionary history, 
selection will have designed psychological adaptations biased in the direction
of the less costly error''.\footnote{%
Such asymmetry in costs is manifest in the so-called \emph{life-dinner}
principle of Richard Dawkins --- 
``The rabbit runs faster than the fox, 
because the rabbit is running for his life
while the fox is only running for his dinner'' --- and 
can exert a strong selection pressure \CITE{Dawkins-Krebs:1979}.
Neuroscientist Joseph LeDoux has a nice way to 
express ``bias by design'' in his book \emph{The Emotional Brain}: 
"It is better to have treated a stick as a snake than not to have responded 
to a possible snake" (\cite[p.166]{LeDoux:1996}).
}
However, the above analysis is performed under the 
so-called ``error management theory'', that is, supposing known the 
probability of the bad outcome.
What happens when the individual 
does not know \emph{a priori} the objective
probability driving the occurence of a bad outcome?

In this paper, we consider the classical problem of a decision maker
(DM) faced with a repeated choice between a
certain option (one known safe payoff) and a risky option
(two known risky payoff values, but unknown probability of each).
Regarding uncertainty, we are thus in the so-called ambiguity setting 
(and not in the risk setting).
Regarding payoffs, we suppose that the three payoffs are ranked in such a way that the safe one 
stands between the two risky ones; thus, the lowest (risky) payoff reflects a
bad outcome. 
We will show that a rational decision maker --- 
in the sense of maximizing expected discounted utility (where the mathematical
expectation involves a prior on the unknown probabilities) ---
can exhibit a behavior displaying ``euphorism'' and status quo biases, as well
as, under suitable conditions, the pessimistic erroneous assessment of the best
objective option 
and an overestimation bias for the probability of the bad outcome.
Thus, in some settings (detailed in the paper), it is quite rational to pay more attention to bad
outcomes than to good ones, and to exaggerate their importance,
even if one aggregates uncertainties by means of a (balanced,
risk-neutral) mathematical expectation, and aggregates payoffs by summation.

It is well-known that the problem we address can be framed as a two-armed bandit
problem, as there are only two decisions, as information is triggered by
decision, as the criterion is intertemporal under unknown probabilities.
In the mathematics and the psychology literature, there is a huge body of work on armed bandit problems,
and on its celebrated solution (when suitable hypothesis are met, see~\CITE{Gittins:1979}) by means of a
dynamic allocation index (Gittins Index Theorem). 
However, this not the route we follow.
Indeed, we revisit this type of problem as an instance of 
optimization problem under imperfect state observation, and we devote a whole part 
to discuss which of our results are robust \wrt\ (with respect to) to assumptions like
stationarity, discounting, finite or infinite horizon.
By doing so, we want to reveal features of optimal strategies that are more
general than those obtained by means of the Gittins index strategy.
Of course, some features --- for instance that the information needed for
optimal decisions can be summarized in a posterior that is updated following
Bayes rule --- are shared with this latter, but they do not depend on the 
mathematical expression of the index.
\bigskip



The paper is organized as follows. 
In Sect.~\ref{sec:learning}, 
we consider the problem of a decision maker
faced with a repeated choice between a
certain option (one known safe payoff) and a risky option
(yielding either a bad or a good outcome, but with unknown probabilities).
We set up a formal mathematical model 
of stochastic sequential decision-making --- under (Bayesian) ambiguity
regarding random sequences of bad and good outcomes (Bernoulli trials) --- 
and we describe an optimal strategy and the behavior of the optimal~DM.
This section contains known results, but with new proofs that make
it  possible to assess the robustness of the findings.
In Sect.~\ref{Killing_three_biases_with_one_stone}, we prove and display
features of the optimal strategy --- optimally designed for a Bayesian
criterion, that is, for a certain (subjective) probability distribution on
sequences of bad and good outcomes --- 
when it is implemented with a Bernoulli process under objective probabilities
(objective environment).
We distinguish two outputs of the optimal strategy ---
estimation of the unknown objective probability of the bad outcome,
assessement of whether the objective environment is prone to risk-taking or not.
When one output is not what it would be were the objective probability
distribution known, we deal with a biais. We summarize in Tables~\ref{tab:Prudence-prone_environment}
and~\ref{tab:Risk-taking-prone_environment} our findings regarding 
consistency or discrepancy \wrt\ what would be optimal in the objective
environment: ``euphorism'' and status quo biases, as well as boundaries and
amplitudes of two effects, the pessimistic erroneous assessment of the best
objective option and the overestimation of the probability of the bad outcome.
In Sect.~\ref{Discussion},
we discuss the cognitive burden of implementing the optimal strategy
(hence the possibility to be an outcome of natural selection),
the robustness of our findings
and possible psychological interpretations,
and we conclude.
Appendix~\ref{Appendix} gathers technical results and proofs.


\section{A mathematical model of repeated decision-making under ambiguity}
\label{sec:learning}

In~\S\ref{Intertemporal_criterion},
we lay out mathematical ingredients to set up a model 
of sequential decision-making under unknown probability,
and formulate an 
expected discounted payoff maximization problem.
In~\S\ref{Structure_of_an_optimal_strategy}, we analyze the structure of an
 optimal strategy, and then 
 we describe the behavior of a decision-maker who adopts such optimal strategy.

This Sect.~\ref{sec:learning} fixes vocabulary, notation and provides the main properties that will be used
to show our main results in Sect.~\ref{Killing_three_biases_with_one_stone}.
The results exposed in this section are not new:  
the structure of an optimal strategy and the induced behavior are
well-known, although they are generally
presented as a consequence of the Gittins Index Theorem, which is not the way we
prove them in Appendix~\ref{Appendix}.
By taking another route for the proofs, we are able to obtain 
(what we think are new) results on i) how the probability of different regimes
in the optimal behavior depends monotonically upon some of the data
(ii) 
which of our results in Sect.~\ref{Killing_three_biases_with_one_stone} 
are robust \wrt\ to assumptions like
stationarity, discounting, finite or infinite horizon
(\S\ref{Discussion_on_the_robusteness_of_the_results_obtained}).

 \subsection{An expected discounted payoff maximization problem}
 \label{Intertemporal_criterion}

In \CITE{Haselton-Nettle:2006}, the following situation is examined.
To reach her/his destination, an individual has two options: 
a short risky route passes through a grassy land --- possibly 
hiding a poisonous snake inflicting serious (though non lethal) pains --- 
whereas a safe route makes a long costly detour. 
Two decisions are possible, with different costs.
If one avoids the grass, one makes a detour that is costly in time,
but one suffers no pain from the (possible) snake.
If one passes through the grass  (``trying'', ``learning'',
``experimenting''), the time spent is shorter but one can 
suffer pain (though not lethal) if the snake is present.
We will illustrate our mathematical setting with this story.

\subsubsection*{Sequential decision-making}

We consider two possible outcomes (states of Nature) --- a bad one~$\bad$
and a good one~$\good$ --- 
that we illustrate by $\bad=$ ``a snake is in the grass'',
and by $\good $ the contrary.
We suppose that, at discrete stages~$t \in \NN$, the~DM makes a decision 
--- either ``avoid'' and be prudent ($\Avoid$) or ``experiment'' and take risks ($\Try$) ---
without knowing in advance the state of Nature occurring at that time
--- either bad ($\bad$) or good ($\good$).
We denote by $t=0, 1, 2\ldots$ the stage corresponding to the 
beginning of the time interval~$[t,t+1[$.
We denote by \( \na{\Avoid,\Try} \) the set of decisions,
and by~\( \control_t \in \na{\Avoid,\Try} \) the action 
taken by the~DM at the beginning of the time interval $[t,t+1[$. 
%
We define the \emph{sample space}
\begin{equation}
 \HISTORY = \{ \bad,\good \}^{\NN^*} = \{ \bad,\good \} \times 
\{ \bad,\good \} \times \cdots \eqfinv
\label{eq:universe} 
\end{equation}
with generic element an infinite sequence $(\uncertain_1,\uncertain_2,\dots)$ of elements 
in \( \{ \bad,\good \} \). 
For $t=1, 2\ldots$, we denote by\footnote{%
We denote random variables by uppercase bold letters.}
\begin{equation}
\va{\Uncertain}_{t} : \HISTORY \to \{ \bad,\good \} \eqsepv
\va{\Uncertain}_{t}(\uncertain_1,\uncertain_2,\dots)= \uncertain_{t} \eqfinv
\label{eq:coordinate_mappings}
\end{equation}
the \emph{state of Nature} realized at the beginning of the time interval $[t,t+1[$,
but that cannot be revealed before the end of~$[t,t+1[$.

\subsubsection*{Strategies}

At the beginning of each time interval~$[t,t+1[$, the~DM can either 
``avoid'' (decision~$\Avoid$) --- in which case the~DM has
no information about the state of Nature --- or 
``experiment'' (decision~$\Try$)-- in
which case the state of Nature \( \va{\Uncertain}_{t+1}\) ($\bad$ or $\good$)
is revealed and experimented, at the end of the time interval~$[t,t+1[$.

We assume that the~DM is not visionary and
learns only from the past:
she/he cannot know the future in advance, neither can the~DM know the state of
Nature ($\bad$ or $\good$) if the~DM decides to avoid. 
We define the \emph{observation sets} at stage $t=0,1,2,3\ldots$ by
\( \OBSERVATION_0 = \{ \partial \} \), 
where \( \partial \) corresponds to no information
(no observation at initial stage $t=0$), and 
\(
\OBSERVATION_t = \{ \bad,\good,\partial \}^{t}
\)
for $t=1,2,3\ldots$
We define the \emph{observation mapping} \( \information : 
\na{\Avoid,\Try} \times  \{ \bad,\good \}
 \to \na{\bad,\good,\partial} \) 
by \( \information(\Try,\bad)=\bad \), \( \information(\Try,\good)=\good \) and
\( \information(\Avoid,\bad)=\information(\Avoid,\good)=\partial \).
Thus, the observation at stage $t=0, 1, 2\ldots$ if the~DM makes decision 
$\control_t \in \na{\Avoid,\Try}$ is 
\( \va{\Observation}_{t+1} = \information\np{\control_t,\va{\Uncertain}_{t+1}}
\).
This case is also known as the  \emph{partial feedback} case,
where foregone payoffs are not revealed.

We allow the~DM to accumulate past observations; therefore
the decision $\control_t$ at stage~$t$ can only be a function of 
\( \va{\Observation}_1, \ldots, \va{\Observation}_{t} \) (the initial decision $\control_0$ is made without 
information).
A \emph{policy at stage~$t$} is a mapping \( \strategy_t : \OBSERVATION_t \to 
\na{\Avoid,\Try} \) that tells the~DM what will be the next action 
in view of past observations.  
A \emph{strategy} \( \strategy \) 
is a sequence \( \strategy = \np{ \strategy_0, \strategy_1, \ldots } \) of policies.
Given 
a strategy~\( \strategy \), decisions and observations
are inductively given by 
\begin{subequations}
  \begin{align}
    \va{\Control}_0 
&= 
\strategy_0 \in \na{\Avoid,\Try} 
\eqfinv
\\
\va{\Observation}_{t+1} 
&= 
\information\np{\va{\Control}_t, \va{\Uncertain}_{t+1}} \in \{ \bad,\good,\partial \} 
\eqsepv \forall t=0,1,2\ldots
\eqfinv
\\
\va{\Control}_t 
&= 
\strategy_t\np{\va{\Observation}_1,\ldots,\va{\Observation}_t} \in
    \na{\Avoid,\Try} 
\eqsepv \forall t=0,1,2\ldots
\eqfinp
  \end{align}
\label{eq:strategy}
\end{subequations}
In the full feedback case, where foregone payoffs are revealed no matter what the
decision made, we have \( \va{\Observation}_{t}= \va{\Uncertain}_{t} \), for
\( t=0,1,2\ldots \).

\subsubsection*{Hypothesized probability}

We introduce the one-dimensional simplex 
\begin{equation}
  \SIMPLEX^1= \bset{ \np{\proba^{\bad},\proba^{\good}} \in \RR^2}%
{\proba^{\bad} \geq 0 \eqsepv \proba^{\good} \geq 0 \eqsepv \proba^{\bad} + \proba^{\good} = 1 }
\eqfinp
\label{eq:simplex}
\end{equation}
The simplex~$\SIMPLEX^1$ is identified with the unit segment~$[0,1]$ by the 
mapping (measurable bijection with measurable inverse)
\( \SIMPLEX^1 \ni \np{\proba^{\bad},\proba^{\good}} \mapsto
\proba^{\bad} \in [0,1] \).
For any \( \np{\proba^{\bad},\proba^{\good}} \in \SIMPLEX^1 \), 
we denote by 
\begin{equation}
\Bernoulli{\proba^{\bad}}{\proba^{\good}} = 
\bigotimes_{t=0}^{\infty}
\bp{ \proba^{\bad} \delta_\bad + \proba^{\good} \delta_\good }
\label{eq:Bernoulli}
\end{equation}
the probability~$\PP$ on the sample space~$\HISTORY$ in~\eqref{eq:universe} 
which makes the 
stochastic process \( (\va{\Uncertain}_1,\va{\Uncertain}_2,\ldots) \) of states of Nature,
as in~\eqref{eq:coordinate_mappings}, a sequence of independent Bernoulli trials 
with marginals given by \( \PP\{ \va{\Uncertain}_t=\bad \} = \proba^{\bad} \) and
\( \PP\{ \va{\Uncertain}_t=\good \} = \proba^{\good} \).

We suppose that the~DM makes the assumption that the 
stochastic process \( (\va{\Uncertain}_1,\va{\Uncertain}_2,\ldots) \) is governed by~\(
\Bernoulli{\proba^{\bad}}{\proba^{\good}} \), but that the~DM  
does not know the probabilities~$\np{\proba^{\bad},\proba^{\good}}$.
Moreover, we suppose that the~DM is a Bayesian who makes the assumption that 
the unknown couple~$(\proba^{\bad},\proba^{\good})$ is a random variable 
with a distribution~$\prior_0$ 
on the one-dimensional simplex~$ \SIMPLEX^1 $ in~\eqref{eq:simplex}.
%
This is why we consider the extended sample space 
$ \SIMPLEX^1 \times\HISTORY = \SIMPLEX^1 \times \{ \bad,\good \}^{\NN^*} $ 
equipped with the probability distribution
\( \prior_0\bp{d\np{\proba^{\bad},\proba^{\good}}} \otimes 
\Bernoulli{\proba^{\bad}}{\proba^{\good}} \),
whose marginal distribution 
on the sample space~$\HISTORY$ in~\eqref{eq:universe}
we denote by~$\PP^{\prior_0}$.
Thus, for any measurable bounded function 
\( \fonctiondeux: \HISTORY \to \RR \), we have that 
\begin{equation}
\EE^{\PP^{\prior_0}} \nc{\fonctiondeux}=
  \int_{ \SIMPLEX^1 } \prior_0\bp{d\np{\proba^{\bad},\proba^{\good}}}
 \EE^{ \Bernoulli{\proba^{\bad}}{\proba^{\good}} } \nc{\fonctiondeux} 
\eqfinp 
\label{eq:marginal_probability_extended_sample_space}
\end{equation}

\subsubsection*{Instantaneous payoffs}

Now, to compare strategies, we will make up a criterion, or an objective
function for the~DM.
In an evolutionary interpretation, payoffs are measured in ``fitness''
unit, for instance ``number of days
alive'' or ``number of days in a reproductive state'', taken as
proxies for the number of offspring. 
The payoffs depend both on the decision and on the state of Nature as
in Table~\ref{tab:costs_BadGood}.
\begin{table}[htbp]
\begin{center}
\begin{tabular}{|c|c|c|}
\hline\hline
& bad state~$\bad$ & 
good state $\good$ 
\\ \hline\hline
  avoid $\Avoid$ & avoidance payoff & avoidance payoff
  \\ 
& $ \utility(\Avoid,\bad)= \util_{\Avoid}$
                       &  $ \utility(\Avoid,\good)=\util_{\Avoid}$
  \\ \hline
experiment $\Try$ & low payoff & high payoff 
  \\
& $ \utility(\Try,\bad)=\util^{\bad}$
                       & $ \utility(\Try,\good)= \util^{\good}$
  \\ \hline
\end{tabular}
\caption{Instant payoffs according to decisions (rows avoid ($\Avoid$) 
or experiment ($\Try$)) 
and states of Nature (columns bad $\bad$ or good $\good$)
\label{tab:costs_BadGood}}
\end{center}
\end{table}

We assume that the payoffs attached to the couple (action, state) in
Table~\ref{tab:costs_BadGood} are ranked as follows:
\begin{equation}
\overbrace{ \utility(\Try,\good)=\util^{\good} }^{\textrm{high payoff}} > 
\underbrace{ \utility(\Avoid,\bad)= \utility(\Avoid,\good)=\util_{\Avoid} }_{\textrm{avoidance (middle) payoff}}  > 
\overbrace{ \utility(\Try,\bad)=\util^{\bad}  }^{\textrm{low payoff}} 
\eqfinp
\label{eq:payoffs}
\end{equation}
In other words, avoiding yields more utility than 
a bad outcome but less than a good one.

\subsubsection*{Intertemporal criterion}

As the payoffs in Table~\ref{tab:costs_BadGood} are measured in ``fitness'',
we suppose that they are cumulative,
like days in a healthy condition or number of offspring. 
This is why we suppose that the~DM can evaluate her/his lifetime performance using 
strategy $ \strategy $ by the discounted intertemporal payoff 
\begin{equation}
\criterion\bp{\strategy,\va{\Uncertain}} = \sum_{t=0}^{+\infty} \discount^{t} 
\utility\np{\va{\Control}_t,\va{\Uncertain}_{t+1}} \; ,
\label{eq:discounted_payoff}
\end{equation}
where \( \va{\Control}_t \) is given by~\eqref{eq:strategy}.
Beyond ``fitness'', our analysis extends to the maximization of any objective function which
can be expressed as an infinite sum over time of discounted payoffs. 

The rationale behind using \emph{discounted} intertemporal payoff is the following.
Suppose that the~DM makes decisions up to a random ultimate stage~$\va{\horizon}$
like, for instance, the DM's lifetime (measured in number of decision stages).
If we suppose that the random variable~$\va{\horizon}$ is 
independent of the randomness in the occurence of a bad and good outcomes,
and follows a (memoryless) Geometric distribution with values in \(
\{0,1,2,3\ldots\} \), then it is easy to establish 
the equality \( \sum_{t=0}^{+\infty} \discount^{t} 
\utility\np{\va{\Control}_t,\va{\Uncertain}_{t+1}} = \EE_{\va{\horizon}} [ \sum_{t=0}^{\va{\horizon}}
\utility\np{\va{\Control}_t,\va{\Uncertain}_{t+1}} ] \),
where the mathematical expectation~\( \EE_{\va{\horizon}} \) is only \wrt\ the
random variable~$\va{\horizon}$. Then, 
we can interpret the discount factor $\discount \in [0,1[$ 
in term of the expected value~$\overline{\va{\horizon}}$ of the random number~$\va{\horizon}$ of
stages during which the~DM has to make decisions, 
by means of the equations $\overline{\va{\horizon}}=\discount/(1-\discount)$ and
\( \discount = \overline{\va{\horizon}} / (\overline{\va{\horizon}}+1)\).
For instance, for an individual making daily decisions during a mean time of one year (resp. fifty years),
we have \( \overline{\va{\horizon}}=365 \) (resp.  \( \overline{\va{\horizon}}=365 \times 50 \)), hence 
\( \discount \approx 0.9972\) (resp.  \( \discount \approx 0.9999 \)).

\subsubsection*{Expected discounted payoff maximization problem}

As the payoff~\eqref{eq:discounted_payoff} is contingent on the
unknown scenario \( \va{\Uncertain} = (\va{\Uncertain}_1,\va{\Uncertain}_2,\ldots)  \), 
it is practically impossible that 
a strategy~\( \strategy \) performs better than another for all scenarios.
We look for an \emph{optimal strategy}~\( \strategy\opt \), solution of 
\begin{equation}
\EE^{\PP^{\prior_0}} \big[\criterion\np{\strategy\opt,\va{\Uncertain}} \big] =
\max_\strategy \EE^{\PP^{\prior_0}} \bc{\criterion\np{\strategy,\va{\Uncertain}}}
\eqfinv
\label{eq:optimization}
\end{equation}
where~$\criterion\np{\strategy,\va{\Uncertain}}$ is given by~\eqref{eq:discounted_payoff},
and the probability~\( \PP^{\prior_0} \),
on the sample space~$\HISTORY$ in~\eqref{eq:universe},
is defined by~\eqref{eq:marginal_probability_extended_sample_space}. 


\subsection{Structure of an optimal strategy and
  behavior of an optimal decision-maker}
\label{Structure_of_an_optimal_strategy}

Here, we analyze the structure of an
 optimal strategy, and then 
 we describe the behavior of a decision-maker who adopts such optimal strategy.

\subsubsubsection{Structure of an optimal strategy}

Let $\Delta(\SIMPLEX^1)$ denote the set of probability distributions
on the simplex~$\SIMPLEX^1$ in~\eqref{eq:simplex}. 
For any \( \InformationState \in \Delta(\SIMPLEX^1) \), we define
\begin{subequations}
  \begin{align}
    \ic{\InformationState}
&=
\int_{\SIMPLEX^1} \np{\proba^{\bad},\proba^{\good}}
\InformationState\bp{d\np{\proba^{\bad},\proba^{\good}}} 
=
\bp{ \ic{\InformationState}^{\bad}, \ic{\InformationState}^{\good} } 
\in \Delta(\SIMPLEX^1) 
\eqfinv
\\
  \ic{\InformationState}^{\bad} 
&=
\int_{\SIMPLEX^1} \proba^{\bad}
  \InformationState\bp{d\np{\proba^{\bad},\proba^{\good}}} \in [0,1]
\eqfinv
\\
 \ic{\InformationState}^{\good}
&=
\int_{\SIMPLEX^1} \proba^{\good} \InformationState\bp{d\np{\proba^{\bad},\proba^{\good}}}  \in [0,1]
\eqfinv 
  \end{align}
that is, the mean of the random variable~\(
\np{\proba^{\bad},\proba^{\good}} \) under probability~$\InformationState$,
and the means of its two components (with \( \ic{\InformationState_{\tau}}^{\bad} +
  \ic{\InformationState_{\tau}}^{\good} =1 \)).
\label{eq:mean_InformationState}
\end{subequations}
%
We also define the two \emph{shift mappings}
\( \theta^{\bad}, \theta^{\good}  : \Delta(\SIMPLEX^1) \to \Delta(\SIMPLEX^1) \)
by
\begin{subequations}
\begin{align}
\np{\theta^{\bad}\InformationState}\bp{d\np{\proba^{\bad},\proba^{\good}}}
&= 
\frac{ \proba^{\bad} }{ \ic{\InformationState}^{\bad} }
\InformationState\bp{d\np{\proba^{\bad},\proba^{\good}}}
\eqfinv
 \\
\np{ \theta^{\good}\InformationState}\bp{d\np{\proba^{\bad},\proba^{\good}}}
&= 
\frac{ \proba^{\good} }{ \ic{\InformationState}^{\good} }
\InformationState\bp{d\np{\proba^{\bad},\proba^{\good}}}
\eqfinp
\end{align}  
\label{eq:mappings_theta}
\end{subequations}
Thus, $\theta^{\bad} \InformationState$ and $\theta^{\good} \InformationState$,
are absolutely continuous with respect to~$\InformationState$.
When \( \ic{\InformationState}^{\bad} = 0 \), that is, when
\( \InformationState=\delta_{\np{0,1}} \), we set 
\( \theta^{\bad}\delta_{\np{0,1}}=\delta_{\np{0,1}}\) and,
when \( \ic{\InformationState}^{\good} = 0 \), that is, when
\( \InformationState=\delta_{\np{1,0}} \), we set 
\( \theta^{\good}\delta_{\np{1,0}}=\delta_{\np{1,0}}\).

As we said at the beginning of this section,
the following result is not new, but we give a proof
(in \S\ref{Proof_of_Propositionpr:Optimal_strategy_under_uncertainty})
that does not rely on the Gittins Index Theorem.

\begin{proposition}
There exists an optimal strategy \( \strategy\opt = \np{ \strategy\opt_0, \strategy\opt_1, \ldots } \) solution of 
the optimization problem~\eqref{eq:optimization}
made of stationary feedback policies of the form 
\begin{equation}
\strategy\opt_t(\va{\Observation}_1,\dots,\va{\Observation}_t) = \widehat{\strategy}(\InformationState_t) 
\eqsepv \forall t=0,1,2\ldots 
\eqfinv
\label{eq:stationary_feedback_policies}
\end{equation}
where 
$\InformationState_t \in \Delta(\SIMPLEX^1)$ is given by 
the dynamical equation
\begin{equation}
\InformationState_0 = \prior_0 \mtext{ and } 
{ \InformationState_{t+1} = 
\dynamics\np{\InformationState_t,\va{\Observation}_{t+1}}=
\begin{cases}
  \InformationState_t & \textrm{if } \va{\Observation}_{t+1} = \partial 
\eqfinv
\\
  \theta^{\bad}\InformationState_t & \textrm{if } \va{\Observation}_{t+1} = \bad 
\eqfinv
\\
  \theta^{\good}\InformationState_t & \textrm{if } \va{\Observation}_{t+1} = \good 
\eqfinp
\end{cases} }  
\label{eq:dynamics}
\end{equation}
Regarding the stationary feedback 
\( \widehat{\strategy} : \Delta(\SIMPLEX^1) \to \{ \Try, \Avoid \} \),
there exists a subset \( \INFORMATIONSTATE_{\Avoid} \subset \Delta(\SIMPLEX^1)
\), and its complementary subset \( \INFORMATIONSTATE_{\Try} =
\Delta(\SIMPLEX^1) \setminus \INFORMATIONSTATE_{\Avoid} \), 
such that 
\begin{itemize}
\item 
\( \widehat{\strategy}(\InformationState) = \Avoid \) 
(that is, select decision ``avoid'') if 
\( \InformationState \in \INFORMATIONSTATE_{\Avoid} \),
\item 
\( \widehat{\strategy}(\InformationState) = \Try \)
(that is, select decision ``experiment'') if 
\( \InformationState \in \INFORMATIONSTATE_{\Try} \).
\end{itemize}
Regarding the complementary subsets \( \INFORMATIONSTATE_{\Avoid} \) and 
\( \INFORMATIONSTATE_{\Try} \), 
there exists a function \( \VALUE : \Delta(\SIMPLEX^1) \to \RR \) 
such that 
\begin{equation}
\InformationState \in \INFORMATIONSTATE_{\Avoid} 
\iff 
\VALUE\np{\InformationState} = 
\frac{\util_{\Avoid}}{1-\discount} 
\eqsepv
\InformationState \in \INFORMATIONSTATE_{\Try}
\iff 
\VALUE\np{\InformationState} > 
\frac{\util_{\Avoid}}{1-\discount} 
\eqfinp 
\label{eq:INFORMATIONSTATE_Try}
\end{equation}
%
%
\label{pr:Optimal_strategy_under_uncertainty}
\end{proposition}

The so-called \emph{information state}~$\InformationState_t \in
\Delta(\SIMPLEX^1)$
is the conditional distribution, with respect to $\va{\Observation}_1,\dots,\va{\Observation}_t$, 
of the first coordinate mapping on $ \SIMPLEX^1 \times\HISTORY $, that is, 
the \emph{posterior} of~\( \np{\proba^{\bad},\proba^{\good}} \) at stage~$t$.

\subsubsubsection{Behavior of an optimal decision-maker}


We call \emph{optimal~DM} 
a decision-maker who adopts the optimal strategy
of Proposition~\ref{pr:Optimal_strategy_under_uncertainty}.
To describe the behavior of an optimal~DM,
we introduce the \emph{first avoidance stage}, or \emph{first prudent stage},
as the random variable defined by 
\begin{equation}
\tau = \inf \bset{ t=0,1,2\ldots }
{ \InformationState_t \in \INFORMATIONSTATE_{\Avoid} }
\eqfinp 
\label{eq:tau}
\end{equation}
In case \( \InformationState_t \not\in \INFORMATIONSTATE_{\Avoid} \) 
for all stages $t=0,1,2\ldots $, 
the convention is $\tau=\inf \emptyset=+\infty$.


As we said at the beginning of this section,
the following result is not new, but we give a proof
(in \S\ref{Proof_of_Propositionpr:Optimal_strategy_behavior})
that does not rely on the Gittins Index Theorem.

\begin{proposition}
The DM that follows the optimal strategy
of Proposition~\ref{pr:Optimal_strategy_under_uncertainty} 
switches at most once
from experimenting to avoiding. More precisely, her/his behavior displays one of
the three following patterns, 
depending on the first avoidance stage~$\tau$ in~\eqref{eq:tau}.
 \begin{enumerate}[a)]
\item 
Infinite risky behavior: \\
if $\tau=+\infty$, that is, if 
\( \InformationState_t \in \INFORMATIONSTATE_{\Try} \) for all stages
$t=0,1,2\ldots $, the optimal~DM always experiments (taking risks),
hence never avoids.
\item 
No risky behavior: \\
if $\tau=0$, that is, if $\prior_0 \not\in \INFORMATIONSTATE_{\Avoid} $
(that is, \( \prior_0 \in \INFORMATIONSTATE_{\Try} \)),
the optimal~DM avoids from the start and, from then on, 
the~DM keeps avoiding (prudence) for all times.
\item 
Finite risky behavior: \\ \label{it:Finite_learning}
if $1 \leq \tau < +\infty$, the optimal~DM 
\begin{itemize}
\item 
experiments (taking risks) from $t=0$ to $\tau-1$, that is, as long as 
\( \InformationState_t \in \INFORMATIONSTATE_{\Try} \), 
\item 
switches to avoiding at stage $t=\tau$, that is, 
as soon as \( \InformationState_t \in \INFORMATIONSTATE_{\Avoid} \),
\item  
from then on, keeps avoiding (prudence) for all times. 
\end{itemize}
\end{enumerate}
\label{pr:Optimal_strategy_behavior}
\end{proposition}


\section{Conditions for biased or accurate assessments}
\label{Killing_three_biases_with_one_stone} 

Now, we show features of the optimal~DM behavior 
that possess interesting psychological
interpretations in terms of human biases:
``euphorism'' and status quo biases in~\S\ref{Status_quo_bias};
possible erroneous assessments of the objective best option and of the objective probabilities
in~\S\ref{Overestimation_of_small_probabilities_bias}.
Contrarily to Sect.~\ref{sec:learning}, 
the results exposed in this Sect.~\ref{Killing_three_biases_with_one_stone} are new.

\subsection{``Euphorism'' and status quo biases}
\label{Status_quo_bias}

Our analysis provides theoretical support to a mix of the 
so-called \emph{status quo bias} ---  
a preference for the current state of affairs
documented in~\cite{Samuelson-Zeckhauser:1988} ---
and to an inclination that we coin ``euphorism'' bias,
related to the ``stay-with-a-winner'' rule ---
if an individual experiments a good outcome, 
it is rational to go on taking risks.

The proof of the following Proposition~\ref{pr:euphorism} can be found in~\ref{Proof_of_Propositionpr:euphorism}. 

\begin{proposition}
  If the optimal~DM experiments a good outcome,
  the DM will go on experimenting (``euphorism'').
 As a consequence, the experimenting phase (in case it exists) of the optimal~DM
 can only stop when a bad outcome materializes: the switch from riskiness to prudence can only be triggered by
the occurrence of a bad outcome. 
 
Therefore, the behavior of the optimal~DM displays at most two consecutive phases
of ``status quo'' --- 
one (possibly empty) of experimenting, that is, taking risks, 
one (possibly empty) of prudence --- with at most one switch;
in particular, once prudent, this is forever.
\label{pr:euphorism}
\end{proposition}


\subsection{Possible erroneous assessments of the objective best option and of the objective probabilities}
\label{Overestimation_of_small_probabilities_bias}

In~\S\ref{Objective_environment}, we formalize what is an objective environment,
with objective best option and probabilities.
By refering to an objective environment, we are able to 
provide a formal definition of a bias in~\S\ref{Formal_definition_of_bias}.
Finally, we study possible erroneous assessments of the objective best option
and of the objective probabilities, for environments prone to prudence
in~\S\ref{The_case_of_environments_prone_to_prudence} and for 
environments prone to risk-taking
in~\S\ref{The_case_of_environments_prone_to_risk-taking}.

\subsubsection{Objective environment}
\label{Objective_environment}

We suppose that Nature produces bad and good
outcomes that are sequences of independent Bernoulli trials governed by 
a given \( \np{\bar\proba^{\bad},\bar\proba^{\good}} \in \SIMPLEX^1 \).
Thus, we equip the sample space~$\HISTORY$ in~\eqref{eq:universe} 
with the probability distribution~$\PP^{\delta_{\np{\bar\proba^{\bad},\bar\proba^{\good}}}}= \Bernoulli{\bar\proba^{\bad}}{\bar\proba^{\good}}$ 
as in~\eqref{eq:Bernoulli}.

\begin{definition}
We call the couple \( \bar\proba=\np{\bar\proba^{\bad},\bar\proba^{\good}} \in \SIMPLEX^1 \) the
\emph{objective} or \emph{true} probabilities.
We call \emph{environment} the triplet \( \np{\discount,\utility,\bar\proba} \)
consisting of
the discount factor $\discount \in [0,1[$,
the payoff function~$\utility$ in Table~\ref{tab:costs_BadGood}
(that is, avoidance payoff~$\util_{\Avoid}$, low payoff~$\util^{\bad}$ 
and high payoff~$\util^{\good}$),
and the objective probabilities
\( \bar\proba=\np{\bar\proba^{\bad},\bar\proba^{\good}} \).

\begin{subequations}
We define the \emph{critical probability $\Critical$} 
by the ratio 
\begin{equation}
\Critical = \frac{\util^{\good}-\util_{\Avoid}}{\util^{\good}-\util^{\bad}}
= \frac{\text{relative payoff of avoidance}}%
{ \text{relative payoff of bad outcome}} \in ]0,1[
  \eqfinv
\label{eq:critical}
\end{equation}
so that we have the equivalence
\begin{equation}
\bar\proba^{\bad} < \Critical   \iff
  \bar\proba^{\bad}\util^{\bad} + \bar\proba^{\good}\util^{\good} >
  \util_{\Avoid}
  \eqfinp
  \label{eq:critical_equivalence}
\end{equation}
When \( \bar\proba^{\bad} < \Critical \) (resp. $\geq$)
or, equivalently, when 
\( \bar\proba^{\bad}\util^{\bad} + \bar\proba^{\good}\util^{\good} > \util_{\Avoid} \) 
(resp. $\leq$), we say that the risky (resp. certain)
option is the \emph{objectively best option} and that 
\emph{the environment is prone to risk-taking} (resp. \emph{prudence}). 
\end{subequations}
\label{de:critical}
\end{definition}
All things being equal, the worse a bad outcome (that is, low payoff of bad
outcome), the lower the critical probability~\eqref{eq:critical}.
When \( \Critical \approx 0 \), the bad outcome is so bad that 
the positive difference between
the payoff of the good outcome and the avoidance payoff is negligible
\wrt\ the positive difference between
the payoff of the good and the bad outcomes;
hence, prudence (avoidance) is the best objective option for most of the
values~$\bar\proba^{\bad}$, since 
\( \bar\proba^{\bad} \geq \Critical \approx 0 \).
When \( \Critical \approx 1 \), the good outcome is so good that
avoiding the bad outcome costs almost as well as suffering it;
taking risks is the best objective option for most of the
values~$\bar\proba^{\bad}$, since \( \bar\proba^{\bad} < \Critical \approx 1 \).

\subsubsection{Formal definition of bias}
\label{Formal_definition_of_bias}

We consider the situation where the optimal~DM 
adopts the strategy
of Proposition~\ref{pr:Optimal_strategy_under_uncertainty},
optimal for a given prior beta distribution~$\prior_0$.
More precisely, let \( n^{\bad}_{0} >0 \) and \( n^{\good}_{0} >0\) be two positive scalars.
We suppose that the distribution~$\prior_0$ is the beta distribution 
$\beta(n^{\bad}_{0},n^{\good}_{0})$ on the simplex~$\SIMPLEX^1$ in~\eqref{eq:simplex},
that is, for any measurable and integrable function 
\( \varphi : \SIMPLEX^1 \to \RR \), we have that 
\begin{equation}
\int_{\SIMPLEX^1} \varphi(\proba^{\bad},\proba^{\good}) 
d\prior_0\np{\proba^{\bad},\proba^{\good}} = \frac{%
\int_0^1 \varphi(\proba,1-\proba) 
\proba^{n^{\bad}_{0}-1} (1-\proba)^{n^{\good}_{0}-1} d\proba}%
{ \int_0^1 \proba^{n^{\bad}_{0}-1} (1-\proba)^{n^{\good}_{0}-1} d\proba} \eqfinp
\label{eq:beta}
\end{equation}

Now, we are equipped to formally define what we call a bias.
On the one hand, the optimal strategy of Proposition~\ref{pr:Optimal_strategy_under_uncertainty},
depends on the discount factor $\discount \in
[0,1[$, on the payoff function~$\utility$ in Table~\ref{tab:costs_BadGood}, 
and on the prior beta distribution~$\prior_0=\beta(n^{\bad}_{0},n^{\good}_{0})$, 
but not on the objective probabilities
\( \np{\bar\proba^{\bad},\bar\proba^{\good}} \).
In other words, optimality is \wrt\ the criterion~\eqref{eq:optimization},
where the mathematical expectation is taken \wrt\ 
the (subjective) probability~\( \PP^{\prior_0} \),
on the sample space~$\HISTORY$ in~\eqref{eq:universe},
as defined by~\eqref{eq:marginal_probability_extended_sample_space}. 
On the other hand, Nature produces bad and good
outcomes that are sequences of independent Bernoulli trials governed by the
objective probabilties \( \np{\bar\proba^{\bad},\bar\proba^{\good}} \in
\SIMPLEX^1 \). Would the DM know \( \np{\bar\proba^{\bad},\bar\proba^{\good}}
\),
she/he would design a strategy maximizing the criterion~\eqref{eq:optimization},
but where the mathematical expectation would be taken \wrt\ 
the (objective) probability~$\PP^{\delta_{\np{\bar\proba^{\bad},\bar\proba^{\good}}}}= \Bernoulli{\bar\proba^{\bad}}{\bar\proba^{\good}}$ 
as in~\eqref{eq:Bernoulli}.
We say that the optimal strategy of
Proposition~\ref{pr:Optimal_strategy_under_uncertainty}
displays a bias when one of its outputs is discrepant with 
what it would be if the objective probability distribution were known.

Thus, the probability distribution~$\PP^{\delta_{\np{\bar\proba^{\bad},\bar\proba^{\good}}}}= \Bernoulli{\bar\proba^{\bad}}{\bar\proba^{\good}}$ 
and the stochastic process \( (\va{\Uncertain}_1,\va{\Uncertain}_2,\ldots) \) 
governed by~\( \Bernoulli{\bar\proba^{\bad}}{\bar\proba^{\good}} \) play the role of a
background reference objective environment against which one can assess the outputs of the optimal
strategy (optimal for~$\prior_0$), and possibly qualify them of biased or not.
We distinguish two outputs of the optimal strategy.
One such output is \( \ic{\InformationState_t}^{\bad} \), 
the mathematical expectation~\eqref{eq:mean_InformationState} of the random variable~$\proba^{\bad}$.
By optimally updating the posterior distribution~\( \InformationState_t \) as in~\eqref{eq:dynamics},
the optimal~DM also updates her/his estimate~\( \ic{\InformationState_t}^{\bad}
\) of the unknown probability~$\bar\proba^{\bad}$ of the bad outcome~$\bad$.
Another output is whether \( \ic{\InformationState_{t}}^{\bad}\util^{\bad} +
\ic{\InformationState_{t}}^{\good}\util^{\good} > \util_{\Avoid} \) or
\( \ic{\InformationState_{t}}^{\bad}\util^{\bad} +
\ic{\InformationState_{t}}^{\good}\util^{\good} \leq \util_{\Avoid} \), that
is, how the~DM assesses whether the environment is prone to risk-taking
or to prudence.

\subsubsection{The case of environments prone to prudence}
\label{The_case_of_environments_prone_to_prudence}

In Table~\ref{tab:Prudence-prone_environment}, we sum up the results of
Proposition~\ref{pr:Optimal_strategy_under_uncertainty_BIAS}
in the case of a prudence-prone environment.
Apart from the ``euphorism'' and status quo biases
(row~2, column~2)
already discussed in~\S\ref{Status_quo_bias},
the optimal~DM displays no bias in the following sense:
she/he makes the accurate assessment that the environment is prone to prudence
(row~5, column~2);
nothing can be said of how~\(
\ic{\InformationState_{\tau}}^{\bad} \) --- the posterior of the bad event when
learning stops ---
is related to the objective probability~\( \bar\proba^{\bad} \)
(hence the empty box in row~4, column~3).

\begin{table}[htbp]
\begin{center}
\begin{tabular}{|c||c|c|}
  \hline\hline
  Environment & Endless risk-taking & Risk-taking \\
  prone to & & then endless prudence  
\\
prudence & $ \tau= +\infty $ & $ 1 \leq \tau < +\infty $ 
\\
\hline\hline
& Behavior \emph{discrepant} 
                                    & Behavior \emph{consistent}
  \\
  &  with the &  with the
\\ 
& feature of the environment & feature of the environment 
\\
\cline{2-3} 
  \( \bar\proba^{\bad} \geq \Critical \)
               & The more likely & The more likely 
\\
$ \iff $ 
& the bad outcome, & the bad outcome, 
\\
\( \bar\proba^{\bad}\util^{\bad} + \bar\proba^{\good}\util^{\good} \) 
& the \emph{lower the probability} & the \emph{higher the probability} 
\\
\( \leq \util_{\Avoid} \) 
& of \emph{discrepant} & of \emph{consistent}
  \\
& endless risk-taking: & endless prudence:
  \\
              & \( \bar\proba^{\bad} \nearrow \implies \)& \( \bar\proba^{\bad} \nearrow \implies \)
\\
    &            \( \PP^{\delta_{\np{\bar\proba^{\bad},\bar\proba^{\good}}}}\ba{ \tau= +\infty } \searrow \) & 
\( \PP^{\delta_{\np{\bar\proba^{\bad},\bar\proba^{\good}}}}\ba{ \tau < +\infty }
                                                                                                                 \nearrow \) 
  \\
  \cline{2-3} 
& \emph{Accurate}  & 
  \\
              &   asymptotic estimation of $\bar\proba^{\bad}$: &
                                          \\
&  \( \lim_{t\to +\infty} \ic{\InformationState_t}^{\bad} = \bar\proba^{\bad} \)&
\\ 
\cline{2-3} 
& Asymptotic  & 
\\ 
& \emph{accurate} assessment & \emph{Accurate} assessment 
\\ 
& that the environment is & that the environment is 
\\
& prone to prudence: & prone to prudence: 
\\
& 
\( \lim_{t\to +\infty} \bp{ \ic{\InformationState_t}^{\bad}\util^{\bad} +
\ic{\InformationState_t}^{\good}\util^{\good} } \) 
& \( \ic{\InformationState_{\tau}}^{\bad}\util^{\bad} +
\ic{\InformationState_{\tau}}^{\good}\util^{\good} \leq \util_{\Avoid} \) 
  \\
              &  \( \leq \util_{\Avoid} \) &
                                          \\
\hline\hline
\end{tabular}
\caption{Optimal behavior in an environment objectively prone to prudence
\label{tab:Prudence-prone_environment}}
\end{center}
 \end{table}

 \subsubsection{The case of environments prone to risk-taking}
\label{The_case_of_environments_prone_to_risk-taking}

 \begin{table}[htbp]
\begin{center}
\begin{tabular}{|c||c|c|}
  \hline\hline
  Environment & Endless risk-taking & Risk-taking \\
  prone to & & then endless prudence  
\\
risk & $ \tau= +\infty $ & $ 1 \leq \tau < +\infty $ 
\\
\hline\hline
                                    & Behavior \emph{consistent}
& Behavior \emph{discrepant} 
  \\
  &  with the &  with the
\\ 
& feature of the environment & feature of the environment 
\\
\cline{2-3} 
  \( \bar\proba^{\bad} < \Critical \)
               & The more unlikely & The more unlikely 
\\
$ \iff $ 
& the bad outcome & the bad outcome 
\\
\( \bar\proba^{\bad}\util^{\bad} + \bar\proba^{\good}\util^{\good} \) 
& the \emph{higher the probability} & the \emph{lower the probability} 
\\
\( > \util_{\Avoid} \) 
& of \emph{consistent} & of \emph{discrepant} 
  \\
& endless risk-taking: & endless prudence: 
  \\
              & \( \bar\proba^{\bad} \searrow \implies \)& \( \bar\proba^{\bad} \searrow \implies \)
\\
    &            \( \PP^{\delta_{\np{\bar\proba^{\bad},\bar\proba^{\good}}}}\ba{ \tau= +\infty } \nearrow \) & 
\( \PP^{\delta_{\np{\bar\proba^{\bad},\bar\proba^{\good}}}}\ba{ \tau < +\infty } \searrow \) 
  \\
  \cline{2-3}
              & & \emph{Vanishing discrepancy:}
                  \\
               & \( \bar\proba^{\bad} \downarrow 0 \implies \)& \( \bar\proba^{\bad} \downarrow 0 \implies \)
\\
    &            \( \PP^{\delta_{\np{\bar\proba^{\bad},\bar\proba^{\good}}}}\ba{ \tau= +\infty } \uparrow 1 \) & 
\( \PP^{\delta_{\np{\bar\proba^{\bad},\bar\proba^{\good}}}}\ba{ \tau < +\infty } \downarrow 0 \) 
  \\
  \cline{2-3}
  & Asymptotic  & 
  \\
  Possible bias 
& \emph{accurate} estimation & \emph{Overestimation} 
  \\
& of the probability~$\bar\proba^{\bad}$ &
of the probability~$\bar\proba^{\bad}$
  \\
              &  of the bad outcome:
                                    & of the bad outcome:
  \\
                &  \( \lim_{t\to +\infty} \ic{\InformationState_t}^{\bad} = \bar\proba^{\bad} \)&
\( \bar\proba^{\bad}  < \Critical \leq \ic{\InformationState_{\tau}}^{\bad}  \)
                                      \\
\cline{2-3} 
& Asymptotic  & 
\\ 
Possible bias 
& \emph{accurate} assessment & \emph{Erroneous} assessment 
\\ 
& that the environment is & that the environment is 
\\
& prone to risk-taking: & prone to prudence: 
\\
& 
\( \lim_{t\to +\infty} \bp{ \ic{\InformationState_t}^{\bad}\util^{\bad} +
\ic{\InformationState_t}^{\good}\util^{\good} } \) 
& \( \ic{\InformationState_{\tau}}^{\bad}\util^{\bad} +
\ic{\InformationState_{\tau}}^{\good}\util^{\good} \leq \util_{\Avoid} \) 
  \\
              & \( > \util_{\Avoid} \) &
                                         \\
\hline\hline
\end{tabular}
\caption{Optimal behavior in an environment objectively prone to risk-taking
\label{tab:Risk-taking-prone_environment}}
\end{center}
 \end{table}

In Table~\ref{tab:Risk-taking-prone_environment}, we sum up the results of
Proposition~\ref{pr:Optimal_strategy_under_uncertainty_BIAS}
in the case of an environment prone to risk-taking, and we
point out two possible biases.
On top of the ``euphorism'' and status quo biases
(row~2, column~3)
already discussed in~\S\ref{Status_quo_bias},
the optimal~DM displays an additional form of pessimism bias.
Indeed, the first (and last) time the optimal~DM stops choosing the risky
option, she/he will erroneously assess that the
environment is prone to prudence (row~6, column~3),
and will overestimate the probability of the bad outcome
(row~5, column~3).
More precisely, we establish the following \emph{Biased Learning Theorem}.
Its proof is a consequence of Proposition~\ref{pr:Optimal_strategy_under_uncertainty_BIAS}
to be found in \S\ref{Optimal_strategy_under_uncertainty_BIAS}.
To our knowledge, these results are new.

 \begin{theorem}[Biased Learning Theorem]
Suppose that the assumptions of Proposition~\ref{pr:Optimal_strategy_under_uncertainty_BIAS} are satisfied
and that \( \prior_0 \in \INFORMATIONSTATE_{\Try} \), so that learning happens,
either infinite or finite.
Suppose that the environment is prone to risk-taking, that is,
(see Definition~\ref{de:critical})
\begin{equation}
  \bar\proba^{\bad} < \Critical
  \eqfinp 
\label{eq:low}
\end{equation}
Then, the optimal~DM (of
Proposition~\ref{pr:Optimal_strategy_under_uncertainty}) 
can only display two behaviors.
\begin{enumerate}
\item 
Either the optimal~DM will experiment forever, and will \emph{accurately} estimate 
asymptotically the objective probability~\( \bar\proba^{\bad} \) of the bad outcome~$\bad$;
this experiment phase happens with a probability
\( \PP^{\delta_{\np{\bar\proba^{\bad},\bar\proba^{\good}}}}\bp{ \InformationState_t \in \INFORMATIONSTATE_{\Try}
\eqsepv \forall t=0,1,2\ldots } \) which goes up to~1 when the objective 
probability~\( \bar\proba^{\bad} \) of the bad outcome~$\bad$ goes down to~0.
In that case, we conclude that the more likely a bad outcome,
the more likely the optimal~DM makes an accurate estimation of its objective probability.
\item 
Or the~DM will experiment during a finite number of stages and then stop
experimenting forever;
this stopping phase happens with a probability
\( \PP^{\delta_{\np{\bar\proba^{\bad},\bar\proba^{\good}}}}\bp{ \exists t=1,2,\ldots \eqsepv \InformationState_t \in 
\INFORMATIONSTATE_{\Avoid} } \) which goes down to~0 when the objective 
probability~\( \bar\proba^{\bad} \) of the bad outcome~$\bad$ goes down to~0.
In that case, we conclude that,
if the objective probability~\( \bar\proba^{\bad} \) of the bad
outcome~$\bad$ is so low that \( \bar\proba^{\bad} \leq \Critical \), then, 
when the experiment phase ends at~$\tau<+\infty$, the optimal~DM will \emph{overerestimate}
the objective probability~\( \bar\proba^{\bad} \) of the bad outcome~$\bad$
because of the inequalities 
\begin{equation}
\bar\proba^{\bad} \leq \Critical \leq \ic{\InformationState_{\tau}}^{\bad}  
\eqfinp
\label{eq:overerestimation}
\end{equation}
However, such an overerestimation happens with a vanishing probability as~\( \bar\proba^{\bad} \downarrow 0 \).
\end{enumerate}
\label{th:Biased_Learning_Theorem}
\end{theorem}

Economists have made the point, coined the \emph{Incomplete Learning Theorem},
that the optimal strategy (to maximize discounted expected utility) does not 
necessarily lead to exactly evaluate the unknown probability 
\CITE{Rothschild:1974,Easley-Kiefer:1988,Brezzi-Lai:2000}.
Thus, optimality does not necessarily lead to perfect accuracy.
Our results point to a \emph{Biased Learning Theorem}, as 
we prove that the departure from accuracy displays 
a bias towards overestimation of the bad outcome
when learning stops. 
However, learning stops (hence overerestimation happens) 
with nonincreasing and vanishing probability as the objective probability of the bad
outcome goes down to zero.

\section{Discussion}
\label{Discussion}

In~\S\ref{Discussion_on_the_implementation_of_an_optimal_strategy}, 
we discuss the possible implementation of an optimal strategy by humans,
and, in~\S\ref{Discussion_on_the_robusteness_of_the_results_obtained}, 
the robustness of our findings, before concluding in~\S\ref{Conclusion}.

\subsection{Possible implementation of an optimal strategy by humans}
\label{Discussion_on_the_implementation_of_an_optimal_strategy}

We discuss the data necessary to design an optimal strategy,
as in Proposition~\ref{pr:Optimal_strategy_under_uncertainty}, 
and the cognitive burden of implementing it,
to see if they are insuperable impediments to its progressive selection during the
course of human evolution.

\subsubsection*{How does the~DM obtain the basic data needed to implement an
  optimal strategy?}

The optimal strategy 
of Proposition~\ref{pr:Optimal_strategy_under_uncertainty}
depends on the discount factor~$\discount$ in~\eqref{eq:discounted_payoff}
and on the payoff function~$\utility$ in Table~\ref{tab:costs_BadGood}
(that is, avoidance payoff~$\util_{\Avoid}$, low payoff~$\util^{\bad}$ 
and high payoff~$\util^{\good}$).
Also, under the assumptions of Proposition~\ref{pr:Optimal_strategy_under_uncertainty_BIAS},
the~DM holds the prior 
beta distribution~$\prior_0=\beta(n^{\bad}_{0},n^{\good}_{0})$ in~\eqref{eq:beta},
where \( n^{\bad}_{0} >0 \) and \( n^{\good}_{0} >0\) are two positive scalars.

We have already discussed, right after Equation~\eqref{eq:discounted_payoff},
how the discount factor~$\discount$ can be related to the mean number of
stages during which the~DM has to make decisions.
We suppose that the~DM knows the avoidance (middle) payoff~$\util_{\Avoid}$.
Then, we suggest a way for the~DM to jointly determine two integers 
\( n^{\bad}_{0} \) and \( n^{\good}_{0} \) for the beta distribution~$\beta(n^{\bad}_{0},n^{\good}_{0})$,
and both the low payoff~$\util^{\bad}$ and the base
payoff~$\util^{\good}$.
The~DM starts by making a risky decision and 
\begin{itemize}
\item either the~DM first enjoys~\( n \) good outcomes~${\good}$ 
--- hence discovering the high payoff~$\util^{\good}$ --- 
before suffering a bad outcome~$\bad$
--- hence discovering the low payoff~$\util^{\bad}$;
in that case, the~DM sets
\( n^{\good}_{0}=n \) and \( n^{\bad}_{0} =1 \);
\item or the~DM first suffers~\( n \) bad outcomes~$\bad$
--- hence discovering the low payoff~$\util^{\bad}$ ---
before enjoying a good outcome~${\good}$
--- hence discovering the high payoff~$\util^{\good}$;
in that case, the~DM sets \( n^{\good}_{0}=1 \) and \( n^{\bad}_{0} =n \).
\end{itemize}
So, at the end of those \( n^{\good}_{0}+ n^{\bad}_{0} \) trials, the~DM 
disposes of the two payoffs~$\util^{\bad}$ and ~$\util^{\good}$,
as well as the two integer parameters \( n^{\bad}_{0} >0 \) and \( n^{\good}_{0} >0 \).

\subsubsection*{What is the stage by stage cognitive load of the optimal~DM?}

We suppose that the~DM holds the prior beta
distribution~$\prior_0=\beta(n^{\bad}_{0},n^{\good}_{0})$, where 
\( n^{\bad}_{0},n^{\good}_{0} \) are integers.
Then, we know from Proposition~\ref{pr:Optimal_strategy_under_uncertainty_BIAS} that the posterior~$\InformationState_t$ 
is the beta distribution \( \beta(n^{\bad}_{0}+N_t^{\bad},n^{\good}_{0}+N_t^{\good}) \)
as in~\eqref{eq:beta_t}.
Thus, at each decision stage~$t$, the cognitive load of the optimal~DM is to
keep track of the two integers \( n^{\bad}_{0}+N_t^{\bad} \) and
\( n^{\good}_{0}+N_t^{\good} \) since, by
Proposition~\ref{pr:Optimal_strategy_under_uncertainty}, 
the optimal decision at stage~$t$ is function of
the posterior~$\InformationState_t$. 

\subsubsection*{How can the~DM make an optimal decision at stage~$t$?}

By Proposition~\ref{pr:Optimal_strategy_under_uncertainty}, 
the~DM has to determine if the current posterior~$\InformationState_t
\in \Delta(\SIMPLEX^1)$ either belongs to the subset \( \INFORMATIONSTATE_{\Avoid} \subset \Delta(\SIMPLEX^1)
\) or to the complementary subset \( \INFORMATIONSTATE_{\Try} =
\Delta(\SIMPLEX^1) \setminus \INFORMATIONSTATE_{\Avoid} \), 
to make an optimal decision at stage~$t$.
As \(
\InformationState_t=\beta(n^{\bad}_{0}+N_t^{\bad},n^{\good}_{0}+N_t^{\good}) \),
the~DM needs to identify in which of two subsets of \( \NN^2 \) --- that is, the
couples of integers corresponding to the subsets
\( \INFORMATIONSTATE_{\Avoid} \) and \( \INFORMATIONSTATE_{\Try} \)
of~$\Delta(\SIMPLEX^1)$ --- does the couple 
\( \np{n^{\bad}_{0}+N_t^{\bad},n^{\good}_{0}+N_t^{\good}} \) belong.

It is hard to say if our mind can design ---  
using the discount factor~$\discount$, the avoidance (middle) payoff~$\util_{\Avoid}$, 
the low payoff~$\util^{\bad}$ and the high payoff~$\util^{\good}$ --- 
and if our brain can hold such a ``mental planar map'' made of couples of
integers.
For instance, for an individual making daily decisions during a mean time of one
year (resp. fifty years), this planar map would consist of \( 365^2=133,225 \) 
(resp. \( (5\times 365)^23 \approx 333~10^6 \) couples of integers labelled with a binary label.
Even if they are not astronomical, these numbers are huge.\footnote{%
We can easily arrive at astronomical figures with general policies.
Indeed, recall that a policy at stage~$t$ is a mapping \( \strategy_t :
\{ \bad,\good,\partial \}^{t} \to \na{\Avoid,\Try} \) that tells the~DM what will be the next action 
in view of past observations.  
Disregarding the irrelevant ``observation''~$\partial$,
a policy at stage~$t$ is a mapping from a set of cardinal~$2^t$ towards a binary
set. If an interval $[t,t+1[$ represents one day, the storage of policies for
one year would be astronomically prohibitive.
This is why the existence of a stationary feedback optimal policy seems good
news. However, the argument of such policy is now an element of \(
\Delta(\SIMPLEX^1) \), that is, a probability distribution on the
one-dimensional simplex. Equivalently, being able to implement the optimal strategy 
of Proposition~\ref{pr:Optimal_strategy_under_uncertainty}
amounts to being able to characterize the two complementary subsets 
\( \INFORMATIONSTATE_{\Avoid} \) and \( \INFORMATIONSTATE_{\Try} \),
which is out of question except with simple rules.}

However, it is possible that a close suboptimal strategy be much more simply encoded by
the following rule: if 
\[
\ic{\InformationState_t}^{\bad} \util^{\bad} + 
\ic{\InformationState_t}^{\good}\util^{\good} =
\frac{n^{\bad}_{0} + N_t^{\bad}}{n^{\bad}_{0} + n^{\good}_{0} + t}\util^{\bad} + 
\frac{n^{\good}_{0} + N_t^{\good}}{n^{\bad}_{0} + n^{\good}_{0} + t}
> \util_{\Avoid}
\] 
then make the risky decision \( \va{\Control}_t = \Try \), else avoid.
A DM adopting this strategy would be more prudent than the optimal~DM
because, by~\eqref{eq:inequality_value_function_Try}, we have that 
\[
\frac{ \ic{\InformationState_t}^{\bad} \util^{\bad} + 
\ic{\InformationState_t}^{\good}\util^{\good} }{1-\discount} > 
\frac{\util_{\Avoid}}{1-\discount} \implies 
\VALUE\np{\InformationState_t} > 
\frac{\util_{\Avoid}}{1-\discount} 
\implies 
\InformationState_t \in \INFORMATIONSTATE_{\Try}
\eqfinp  
\]

\subsection{Robusteness of the results obtained}
\label{Discussion_on_the_robusteness_of_the_results_obtained}

We discuss which of our results are robust \wrt\ to assumptions like
stationarity (of the primitive random variables, of the payoffs),
discounting, finite or infinite horizon.

\subsubsection*{``Euphorism'' and status quo biases}

The property that, if the optimal~DM experiments a good outcome,
she/he will go on taking risks is a consequence of the ``stay-with-a-winner''
property~\eqref{eq:stay-with-a-winner}:
the value function cannot decrease when the posterior changes following a 
good outcome. 
Screening the proof of~\eqref{eq:stay-with-a-winner} shows that this property
only depends on the ranking~\eqref{eq:payoffs} of payoffs, hence that our finding
(``euphorism'' bias)
is robust \wrt\ nonstationarity (as long as it does not change the ranking), absence of discounting,
and finite or infinite horizon.

The property that, if one selects the prudent decision once,
one will no longer make risky decisions afterwards is a consequence of both
the stopping of posterior updating and of the stationarity of the avoidance
domain~\( \INFORMATIONSTATE_{\Avoid} \).
We discuss both of them.
The property that the posterior is a sufficient information state for optimization
is quite robust, as it holds true under nonstationarity, absence of discounting,
and finite or infinite horizon (\cite[Chap.~10]{Bertsekas-Shreve:1996}).
The stopping of posterior updating follows from the information structure, 
as avoidance freezes observation, hence is robust.
However, these are stationarity and infinite horizon that lead to status quo.
Indeed, were the avoidance domain~\( \INFORMATIONSTATE_{\Avoid} \) dependent on
the stage~$t$ that we could no longer conclude to status quo.
Thus, the status quo bias is less robust than the ``euphorism'' bias.

\subsubsection*{Pessimistic erroneous assessment of the environment
and overestimation bias for the probability of the bad outcome}

The property that the probability of a bad outcome is 
overestimated when the risky phase stops
(hence the erroneous assessment that the environment is prone to prudence)
comes from the
inequality~\eqref{eq:inequality_value_function_Try}, itself 
a consequence of stationarity, discounting and infinite horizon.
In this sense, it is less robust than the ``euphorism'' bias.

\subsection{Conclusion}
\label{Conclusion}

Our model and analysis show that certain biases
can be the product of rational behavior, 
here in the sense of maximizing expected discounted utility
(that is, being risk neutral)  with learning.
Indeed, our theoretical results provide support to
``euphorism'' and status quo biases, as well as, under narrow boundary
conditions, the pessimistic erroneous assessment of the best objective option
and an overestimation bias for the probability of bad outcomes.
In particular, we have shown a Biased Learning Theorem that 
provides rational ground for the human bias that consists in
attributing to bad outcomes an importance 
larger than their statistical occurrence.
Let us dwell on this point. 

In many situations, probabilities are not known but learnt.
The 2011 nuclear accident in Japan has led many countries 
to stop nuclear energy. This sharp switch may be interpreted as the stopping 
of an experiment phase where the probability of nuclear accidents 
has been progressively learnt. 
In financial economics, the equity premium puzzle comes from the observation 
that bonds are underrepresented in portfolios, despite the empirical fact that 
stocks have outperformed bonds over the last century in the USA 
by a large margin \CITE{Mehra-Prescott:1985}.
However, this analysis is done \emph{ex post} under risk, while 
decision-makers make their decisions day by day under uncertainty, 
and sequentially learn about the probability of stocks losses. \emph{Ex ante}, 
the underrepresentation of bonds can be enlightened by the Biased Learning 
Theorem: the (small) probability of (large) bonds losses is
overestimated with respect to their statistical occurrence.

To end up, our results point to the fact that 
overestimation depends upon relative payoffs 
by the formula~\eqref{eq:critical}.
This property could possibly be tested in experiments.   
\bigskip

\paragraph*{Acknowledgments}


The author thanks the following colleagues for their comments:
Daniel Nettle (Newcastle University), 
Nicolas Treich (Toulouse School of Economics) and
Christopher Costello\footnote{%
I thank Christopher Costello for suggesting the title of the paper.}
 (University of California Santa Barbara); 
Jean-Marc
Tallon, Alain Chateauneuf, Michelle Cohen, Jean-Marc Bonnisseau 
and the organizers of the Economic Theory Workshop of 
Paris School of Economics (Friday 4 November 2011);
John Tooby, Andrew W. Delton, Max Krasnow and 
the organizers of the seminar of the Center for Evolutionary Psychology,
University of California Santa Barbara (Friday 18 November 2011);
Arthur J. Robson and the organizers of the Economics seminar 
at Simon Fraser University (Tuesday 21 October 2014);
Pierre Courtois, Nicolas Querou, Rapha\"{e}l Soubeyran 
 and the organizers of the seminar of Lameta, Montpellier
(Monday 3 October 2016);
Khalil Helioui, Geoffrey Barrows, 
Jean-Pierre Ponssard, Guillaume Hollard, 
Guy Meunier and the organizers of the Sustainable Economic and Financial
Development Seminar at \'Ecole Polytechnique (Tuesday 17 January 2017);
Jeanne Bovet and Luke Glowacki, organizers of the 
Tuesday Lunch at Institute for Advanced Study in Toulouse (Tuesday 4 July 2017).


This research did not receive any specific grant from funding agencies in the public, commercial, or not-for-profit sectors.

\appendix

\section{Technical results and proofs}
\label{Appendix}

\subsection{Proof of Proposition~\ref{pr:Optimal_strategy_under_uncertainty}}
\label{Proof_of_Propositionpr:Optimal_strategy_under_uncertainty}


\begin{proof}
We follow the approach of \cite[Chap.~10]{Bertsekas-Shreve:1996} 
for the analysis of imperfect state information models.

\begin{subequations}
The infinite horizon imperfect state information stochastic optimization
problem~\eqref{eq:optimization}-\eqref{eq:discounted_payoff}-\eqref{eq:strategy}
can be written as\footnote{%
We do not detail over which possible solutions the two
suprema are taken. The left hand side supremum is \wrt~\eqref{eq:strategy},
whereas the right hand side supremum is \wrt\ suitable stochastic kernels
deduced from the information structure below.}
\begin{equation}
  \sup \EE^{\PP^{\prior_0}} \Bc{ \sum_{t=0}^{+\infty} \discount^{t} 
\utility\np{\va{\Control}_t,\va{\Uncertain}_{t+1}} }
=  \sup \int_{\SIMPLEX^1} \Bc{ \sum_{t=0}^{+\infty} \discount^{t}
  \coutint\np{\state_t,\control_t} } \prior_0\np{d\state_0}
\end{equation}
where we have introduced the one-dimensional simplex~$ \SIMPLEX^1 $
in~\eqref{eq:simplex} as state space, and the state dynamics
\begin{equation}
  \state_0 = \np{\proba^{\bad},\proba^{\good}} \in \SIMPLEX^1 
\eqsepv
\state_{t+1} = \state_{t} \eqsepv \forall t=0,1,2\ldots 
\eqfinv
\end{equation}
or, equivalently, 
the state space~\( \SIMPLEX^1 \) and the state transition kernels
\begin{equation}
  \kernel{\State}{d\state}{\state,\control}
=\kernel{\State}{d\np{\proba^{\bad},\proba^{\good}}}{\np{\proba^{\bad},\proba^{\good}},\control}
=\delta_{\state}=\delta_{\np{\proba^{\bad},\proba^{\good}}}
\eqfinv
\end{equation}
the control space~\( \na{\Avoid,\Try}  \) and the controls
\begin{equation}
  \control_t \in \na{\Avoid,\Try}  \eqsepv \forall t=0,1,2\ldots 
\eqfinv
\end{equation}
the one-stage payoff 
\begin{equation}
  \begin{split}
  \coutint\np{\state,\control}=  \coutint\bp{\np{\proba^{\bad},\proba^{\good}},\control}=
\proba^{\bad}\utility\np{\control,\bad} +
\proba^{\good}\utility\np{\control,\good}
\\=
  \begin{cases}
    \util_\Avoid & \text{ if } \control=\Avoid
\eqfinv
\\
\proba^{\bad}\util^{\bad} + \proba^{\good}\util^{\good}
& \text{ if } \control=\Try
\eqfinv
  \end{cases}
  \end{split}
\label{eq:one-stage_payoff}
\end{equation}
and the observation space~\( \na{\bad,\good,\partial} \) 
and the observation stochastic kernel
\begin{equation}
  \begin{split}
  \kernel{\Observation}{d\observation}{\state,\control}=
\kernel{\Observation}{d\observation}{\np{\proba^{\bad},\proba^{\good}},\control}
\\=
  \begin{cases}
    \delta_{\partial}\np{d\observation} & \text{ if } \control=\Avoid
\eqfinv
\\
\proba^{\bad}\delta_{\bad}\np{d\observation} 
+ \proba^{\good}\delta_{\good}\np{d\observation}
& \text{ if } \control=\Try
\eqfinp
  \end{cases}    
  \end{split}
\label{eq:observation_stochastic_kernel}
\end{equation}
The stochastic kernel 
\begin{equation}
  \kernel{\State}{d\state}{\InformationState,\control,\observation} =
\begin{cases}
  \InformationState\np{d\state} & \textrm{if } 
\control=\Avoid \textrm{ and } \observation= \partial 
\eqfinv
\\
  \np{\theta^{\bad}\InformationState}\np{d\state} & \textrm{if } 
\control=\Try \textrm{ and } \observation= \bad 
\eqfinv
\\
  \np{\theta^{\good}\InformationState}\np{d\state} & \textrm{if } 
\control=\Try \textrm{ and } \observation= \good 
\eqfinp
\end{cases} 
\label{eq:induced_stochastic_kernel}
\end{equation}
\end{subequations}
satisfies \cite[Lemma~10.3]{Bertsekas-Shreve:1996}, because it can be checked
that, for any measurable subset \( \SIMPLEX \subset \SIMPLEX^1 \) and 
subset \( C \subset \na{\bad,\good,\partial} \), and 
any \( \control \in \na{\Avoid,\Try} \), one has that 
\begin{equation}
\int_{\SIMPLEX} \kernel{\Observation}{C}{\state,\control}\InformationState\np{d\state}=
\int_{\SIMPLEX^1} \Bc{ \int_{C}
\kernel{\State}{\SIMPLEX}{\InformationState,\control,\observation} 
\kernel{\Observation}{d\observation}{\state,\control} }
\InformationState\np{d\state}
\eqfinp
\label{eq:Lemma10.3_Bertsekas-Shreve:1996}
\end{equation}
Indeed, for \( \control=\Avoid \), Equation~\eqref{eq:Lemma10.3_Bertsekas-Shreve:1996}
is satisfied because
\begin{align*}
&
\int_{\SIMPLEX^1} \Bc{ \int_{C}
\kernel{\State}{\SIMPLEX}{\InformationState,\Avoid,\observation} 
\kernel{\Observation}{d\observation}{\state,\Avoid} }
  \InformationState\np{d\state}
\\
&=
\int_{\SIMPLEX^1} \Bc{ \int_{C}
\InformationState\np{\SIMPLEX}\delta_{\partial}\np{d\observation}}
  \InformationState\np{d\state}
\intertext{by the expressions~\eqref{eq:induced_stochastic_kernel}
for \( \kernel{\State}{d\state}{\InformationState,\Avoid,\observation} \)
and~\eqref{eq:observation_stochastic_kernel} for
$\kernel{\Observation}{d\observation}{\state,\Avoid}$}
&=\delta_{\partial}\np{C}\InformationState\np{\SIMPLEX}
= \int_{\SIMPLEX}\delta_{\partial}\np{C}\InformationState\np{d\state}
\\
&=\int_{\SIMPLEX}
  \kernel{\Observation}{C}{\state,\Avoid}\InformationState\np{d\state}
\tag{by the expression~\eqref{eq:observation_stochastic_kernel} for
$\kernel{\Observation}{d\observation}{\state,\Avoid}$}
\eqfinp
\end{align*}
For \( \control=\Try \), we show that Equation~\eqref{eq:Lemma10.3_Bertsekas-Shreve:1996}
is satisfied for \( C= \na{\bad},\na{\good} ,\na{\partial} \).
For \( C=\na{\partial} \), both sides of the
Equation~\eqref{eq:Lemma10.3_Bertsekas-Shreve:1996}
are zero as \( \kernel{\Observation}{\na{\partial}}{\state,\Try}= 0 \)
by the expression~\eqref{eq:observation_stochastic_kernel} for
$\kernel{\Observation}{d\observation}{\state,\Try}$.
For \( C=\na{\bad} \), we calculate
\begin{align*}
&
\int_{\SIMPLEX^1} \Bc{ \int_{\na{\bad}}
\kernel{\State}{\SIMPLEX}{\InformationState,\Try,\observation} 
\kernel{\Observation}{d\observation}{\state,\Try} }
  \InformationState\np{d\state}
\\
&=
\int_{\SIMPLEX^1} 
\kernel{\State}{\SIMPLEX}{\InformationState,\Try,\bad}
\kernel{\Observation}{\na{\bad}}{\state,\Try} 
  \InformationState\np{d\state}
\\
&=
\int_{\SIMPLEX^1} \np{\theta^{\bad}\InformationState}\np{\SIMPLEX} 
\, \proba^{\bad} \InformationState\bp{d\np{\proba^{\bad},\proba^{\good}}}
\intertext{by the expressions~\eqref{eq:induced_stochastic_kernel}
for \( \kernel{\State}{d\state}{\InformationState,\Try,\observation} \)
and~\eqref{eq:observation_stochastic_kernel} for
$\kernel{\Observation}{d\observation}{\state,\Try}$}
&=
\frac{ 1 }{ \int_{\SIMPLEX^1} \proba^{\bad} \InformationState\bp{d\np{\proba^{\bad},\proba^{\good}}} } 
\int_{\SIMPLEX^1} 
\bc{ \int_{\SIMPLEX} \probabis^{\bad} \InformationState\bp{d\np{\probabis^{\bad},\probabis^{\good}}} }
\proba^{\bad} \InformationState\bp{d\np{\proba^{\bad},\proba^{\good}}}
\tag{by the expression~\eqref{eq:mappings_theta} of 
\( \theta^{\bad}\InformationState \)}
\\
&=
\int_{\SIMPLEX} \probabis^{\bad}
  \InformationState\bp{d\np{\probabis^{\bad},\probabis^{\good}}} 
\\
&=
\int_{\SIMPLEX}
  \kernel{\Observation}{\na{\bad}}{\np{\proba^{\bad},\proba^{\good}},\Try}
\InformationState\bp{d\np{\proba^{\bad},\proba^{\good}}}
\tag{by the expression~\eqref{eq:observation_stochastic_kernel} for
$\kernel{\Observation}{d\observation}{\np{\proba^{\bad},\proba^{\good}},\Try}$}
\\
&=
\int_{\SIMPLEX}
  \kernel{\Observation}{\na{\bad}}{\state,\Try}\InformationState\np{d\state}
\eqfinp 
\end{align*}
For \( C=\na{\good} \), we obtain Equation~\eqref{eq:Lemma10.3_Bertsekas-Shreve:1996}
in the same way.

By \cite[Propositions~10.5 and 10.6]{Bertsekas-Shreve:1996}, we conclude that 
the imperfect state information model can be reduced to a perfect state one,
with new information state $\InformationState \in \Delta(\SIMPLEX^1)$,
new information state transition kernels
\begin{equation}
  \kernel{\INFORMATIONSTATE}{d\InformationState}{\InformationState,\control}
=
  \begin{cases}
    \InformationState & \text{ if } \control=\Avoid
\eqfinv
\\
\ic{\InformationState}^{\bad} \delta_{\theta^{\bad}\InformationState}
+
\ic{\InformationState}^{\good} \delta_{\theta^{\good}\InformationState}
& \text{ if } \control=\Try
\eqfinv
  \end{cases}
\label{eq:new_information_state_stochastic_kernel}
\end{equation}
where $\ic{\InformationState}^{\bad}$ and $\ic{\InformationState}^{\good}$
have been defined in~\eqref{eq:mean_InformationState},
and new one-stage payoff 
\begin{align}
\tilde\coutint\np{\InformationState,\control}
&=
\int_{\SIMPLEX^1} \coutint\np{\state,\control} \InformationState\np{d\state}
\label{eq:new_one-stage_payoff}
\\
&= 
\int_{\SIMPLEX^1} \bc{ \proba^{\bad}\utility\np{\control,\bad} +
\proba^{\good}\utility\np{\control,\good} }
\InformationState\bp{d\np{\proba^{\bad},\proba^{\good}}} 
\tag{by~\eqref{eq:one-stage_payoff}}
 \\
&= 
  \begin{cases}
    \util_\Avoid & \text{ if } \control=\Avoid
\eqfinv
\\
\ic{\InformationState}^{\bad} \util^{\bad} + 
\ic{\InformationState}^{\good} \util^{\good}
& \text{ if } \control=\Try
\eqfinp
  \end{cases}
\nonumber
\end{align}
The value function \( \VALUE : \Delta(\SIMPLEX^1) \to \RR \) given by
\begin{equation}
  \VALUE\np{\InformationState} =
  \sup \EE^{\PP^{\InformationState}} \Bc{ \sum_{t=0}^{+\infty} \discount^{t} 
\utility\np{\va{\Control}_t,\va{\Uncertain}_{t+1}} }\\
=  \sup \int_{\SIMPLEX^1} \Bc{ \sum_{t=0}^{+\infty} \discount^{t}
  \coutint\np{\state_t,\control_t} } \InformationState\np{d\state_0}
\label{eq:value_function}
\end{equation}
satisfies, by \cite[Proposition~9.8]{Bertsekas-Shreve:1996},
the dynamic programming equation
\begin{equation}
\VALUE\np{\InformationState} = \max_{ \control \in \na{\Avoid,\Try} }
\bgp{ \tilde\coutint\np{\InformationState,\control} + 
\int_{\SIMPLEX^1} \kernel{\INFORMATIONSTATE}{d\InformationState'}{\InformationState,\control}
\VALUE\np{\InformationState'} }
\eqfinv 
\label{eq:Bellman}
\end{equation}
that is, by~\eqref{eq:new_one-stage_payoff}
and \eqref{eq:new_information_state_stochastic_kernel},
\begin{equation}
\VALUE\np{\InformationState} = \max 
\Ba{ 
\util_{\Avoid} + \discount\VALUE\np{\InformationState},
\ic{\InformationState}^{\bad} 
\bp{ \util^{\bad} + \discount\VALUE\np{\theta^{\bad}\InformationState} }
+ 
\ic{\InformationState}^{\good} 
\bp{ \util^{\good} + \discount\VALUE\np{\theta^{\good}\InformationState} 
} } 
\eqfinp
\label{eq:Bellmanbis}
\end{equation}
By definition~\eqref{eq:value_function} of
the value function~\( \VALUE : \Delta(\SIMPLEX^1) \to \RR \), we have that 
\begin{subequations}
  \begin{align}
  \VALUE\np{\InformationState} 
& \geq 
 \sum_{t=0}^{+\infty} \discount^{t} 
\int_{\SIMPLEX^1} \coutint\np{\state_t,\Avoid} \InformationState\np{d\state_0}
= 
 \sum_{t=0}^{+\infty} \discount^{t} \util_{\Avoid} 
=
\frac{\util_{\Avoid}}{1-\discount}
\label{eq:inequality_value_function_Avoid}
\intertext{by applying the open-loop strategy \( \strategy_t =\Avoid \)
for all~$t$, and we also have that}
  \VALUE\np{\InformationState} 
& \geq 
\frac{ \ic{\InformationState}^{\bad} \util^{\bad} + 
\ic{\InformationState}^{\good}\util^{\good} }{1-\discount}
\eqfinp
\label{eq:inequality_value_function_Try}
  \end{align}
\end{subequations}
Indeed, we have that 
\begin{align*}
    \VALUE\np{\InformationState} 
& \geq 
\sum_{t=0}^{+\infty} \discount^{t} \int_{\SIMPLEX^1} 
\coutint\np{\state_t,\Try} \InformationState\np{d\state_0}
\tag{by applying the open-loop strategy \( \strategy_t =\Try \)
for all~$t$}
\\
&=\sum_{t=0}^{+\infty} \discount^{t}  \int_{\SIMPLEX^1} 
\bc{ \proba^{\bad}\util^{\bad} + \proba^{\good}\util^{\good} }
\InformationState\bp{d\np{\proba^{\bad},\proba^{\good}}}
\tag{by~\eqref{eq:one-stage_payoff}} 
\\
&=
\frac{ \ic{\InformationState}^{\bad} \util^{\bad} + 
\ic{\InformationState}^{\good}\util^{\good} }{1-\discount}
\tag{by~\eqref{eq:mean_InformationState}.}
\end{align*}
The existence of the proposed stationary optimal policy is given by 
\cite[Proposition~9.12, Corollary~9.12.1, Corollary~9.17.1]{Bertsekas-Shreve:1996}:
depending whether the maximum in the dynamic programming
equation~\eqref{eq:Bellman} is achieved for \( \control=\Avoid \)
or for \( \control=\Try \), we select an optimal strategy accordingly.
\begin{subequations}
This is why we define the subset \( \INFORMATIONSTATE_{\Avoid} \subset \Delta(\SIMPLEX^1) \)
by 
\begin{equation*}
\InformationState \in \INFORMATIONSTATE_{\Avoid} \iff 
\VALUE\np{\InformationState} = 
\util_{\Avoid} + \discount\VALUE\np{\InformationState} 
\iff 
\VALUE\np{\InformationState} = 
\frac{\util_{\Avoid}}{1-\discount} 
\eqfinv
\end{equation*}
which gives the first part of~\eqref{eq:INFORMATIONSTATE_Try}.
From the inequality~\eqref{eq:inequality_value_function_Avoid}, 
we deduce the second part of~\eqref{eq:INFORMATIONSTATE_Try}:
 \begin{equation*}
\InformationState \not\in \INFORMATIONSTATE_{\Avoid} \iff 
\InformationState \in \INFORMATIONSTATE_{\Try}  \iff 
\VALUE\np{\InformationState} > 
\frac{\util_{\Avoid}}{1-\discount} 
\eqfinp 
\end{equation*}
\end{subequations}
\medskip

This ends the proof.
\end{proof}

\subsection{Proof of Proposition~\ref{pr:Optimal_strategy_behavior}}
\label{Proof_of_Propositionpr:Optimal_strategy_behavior}

\begin{proof} 
\quad
  \begin{enumerate}[a)]
  \item 
By Proposition~\ref{pr:Optimal_strategy_under_uncertainty}, 
when $\tau=+\infty$ --- that is, when 
\( \InformationState_t \in \INFORMATIONSTATE_{\Try} \) 
for all stages~$t$ ---
it is optimal to select decision~$\Try$ and to experiment forever.
\item 
By Proposition~\ref{pr:Optimal_strategy_under_uncertainty}, when $\tau=0$ 
--- that is, when \( \prior_0 \in \INFORMATIONSTATE_{\Avoid} \) --- 
it is optimal to select decision~$\Avoid$ and to avoid for all times.
Indeed, once the optimal~DM avoids, the~DM does not observe the random outcomes,
hence the~DM no longer updates the posterior 
$\InformationState_t$ in~\eqref{eq:dynamics}, 
so that the~DM keeps avoiding.
\item 
When $1 \leq \tau<+\infty$, we have, by definition~\eqref{eq:tau} of 
the first avoidance stage~$\tau$, 
   \begin{itemize}
   \item 
\( \InformationState_t \in \INFORMATIONSTATE_{\Try} \) 
for stages $t=0$ up to $\tau -1$;
hence, by Proposition~\ref{pr:Optimal_strategy_under_uncertainty}, 
it is optimal to select decision $\Try$ and experiment 
from stages $t=0$ up to $\tau -1$;
\item 
\( \InformationState_t \in \INFORMATIONSTATE_{\Avoid} \)
 for stages $t=\tau$ up to $+\infty$;
hence, by Proposition~\ref{pr:Optimal_strategy_under_uncertainty}, 
it is optimal to select decision $\Avoid$ and avoid
 for stages $t=\tau$ up to $+\infty$.
Indeed, once the optimal~DM avoids, the~DM does not observe the random outcomes,
hence the~DM no longer updates the posterior 
$\InformationState_t$ in~\eqref{eq:dynamics}, 
so that the~DM keeps avoiding.
   \end{itemize}
\end{enumerate}
This ends the proof.
\end{proof}

\subsection{Proof of Proposition~\ref{pr:euphorism}}
\label{Proof_of_Propositionpr:euphorism}

\begin{proof}
  First, we prove that, if the optimal~DM experiments a good outcome,
  the DM will go on experimenting.
  
Suppose that, at stage~$t$ the optimal~DM is experimenting.
We will show that, if a good outcome~$\good$ materializes at the end
of the interval~$[t,t+1[$ (that is, if $\va{\Observation}_{t+1}=\good$), then necessarily 
the optimal~DM goes on experimenting at stage~$t+1$. 
In what follows, the value function~\( \VALUE : \Delta(\SIMPLEX^1) \to \RR \)
has been introduced in
Proposition~\ref{pr:Optimal_strategy_under_uncertainty},
and is defined in~\eqref{eq:value_function}.
We have that 
\begin{align*}
  \VALUE\np{ \InformationState_{t+1} }
&= 
\VALUE\np{ \theta^{\good} \InformationState_t }
\intertext{because, as we supposed that $\va{\Observation}_{t+1}=\good$, we have that 
\( \InformationState_{t+1} = \theta^{\good} \InformationState_t \) 
by the dynamics~\eqref{eq:dynamics}}
& \geq 
\VALUE( \InformationState_t ) 
\tag{by the property \( \VALUE \circ \theta^{\good} \geq \VALUE \),
shown afterward}
\\
& >
\frac{\util_{\Avoid}}{1-\discount} 
\end{align*} 
by~\eqref{eq:INFORMATIONSTATE_Try}, 
because, as we supposed that the optimal~DM is experimenting at stage~$t$, we have that 
\( \InformationState_t \in \INFORMATIONSTATE_{\Try} \).
Thus, we have obtained that 
\( \VALUE( \InformationState_{t+1} ) > \frac{\util_{\Avoid}}{1-\discount} \).
By~\eqref{eq:INFORMATIONSTATE_Try}, we conclude that 
the optimal~DM goes on experimenting at stage~$t+1$ 
by Proposition~\ref{pr:Optimal_strategy_under_uncertainty}. 

We now prove that the value function~\eqref{eq:value_function} 
has the property 
\begin{equation}
  \VALUE \circ \theta^{\good} \geq \VALUE 
\eqfinv
\label{eq:stay-with-a-winner}
\end{equation} 
that is, the value function cannot decrease when the posterior changes following a 
good outcome. We will use this ``stay-with-a-winner'' property when we discuss the robustness of our findings
in~\S\ref{Discussion_on_the_robusteness_of_the_results_obtained}.

Before that, we recall that two random variables~$\va{C}$ and~$\va{D}$, 
defined on a probability space~$\epro$, are said to be \emph{comonotonic} when we
have that
\( \bp{ \va{C}(\omega)-\va{C}(\omega') }\bp{ \va{D}(\omega)-\va{D}(\omega') }
\geq 0 \), for any \( \np{\omega,\omega'} \in \Omega^2 \).
In that case, it is easily shown that
\( \EE \bc{ \va{C} \va{D} } \geq \EE \bc{ \va{C} } \EE \bc{ \va{D} } \),
when $\va{C}$ and~$\va{D}$ are square integrable.

By definition~\eqref{eq:value_function} of the value function~$\VALUE$,
to prove~\eqref{eq:stay-with-a-winner}
it suffices to show that 
\[
\int_{\SIMPLEX^1} \coutint\np{\state,\control}
\np{\theta^{\good}\InformationState}\np{d\state}
\geq
\int_{\SIMPLEX^1} \coutint\np{\state,\control} \InformationState\np{d\state}
\eqsepv \forall  \control \in \na{\Avoid,\Try} 
\eqfinp
\]
This is obvious for \( \control=\Avoid \) since 
\( \coutint\np{\state,\Avoid}= \util_\Avoid \) by~\eqref{eq:one-stage_payoff}.
For \( \control=\Try \), we have that 
\begin{align*}
  \int_{\SIMPLEX^1} \coutint\np{\state,\Try}
\np{\theta^{\good}\InformationState}\np{d\state}
&=
  \int_{\SIMPLEX^1} \bc{ \proba^{\bad}\util^{\bad} + 
\proba^{\good}\util^{\good} } \frac{ \proba^{\good} }{ \ic{\InformationState}^{\good} }
\InformationState\bp{d\np{\proba^{\bad},\proba^{\good}}}
\tag{by~\eqref{eq:one-stage_payoff} and \eqref{eq:mappings_theta}}
\\
& \geq
\frac{ 1 }{ \ic{\InformationState}^{\good} }
\int_{\SIMPLEX^1} \bc{ \proba^{\bad}\util^{\bad} + 
\proba^{\good}\util^{\good} }
  \InformationState\bp{d\np{\proba^{\bad},\proba^{\good}}}
\int_{\SIMPLEX^1} \proba^{\good} 
  \InformationState\bp{d\np{\proba^{\bad},\proba^{\good}}}
\intertext{because the random variables \( \va{C}: \SIMPLEX^1 \ni \np{\proba^{\bad},\proba^{\good}}
\mapsto \proba^{\bad}\util^{\bad} + \proba^{\good}\util^{\good} \)
and \( \va{D}: \SIMPLEX^1 \ni \np{\proba^{\bad},\proba^{\good}}
  \mapsto \proba^{\good} \) are comonotonic, since the function 
\( \proba^{\good}  \mapsto  \proba^{\bad}\util^{\bad} +
  \proba^{\good}\util^{\good}
= \proba^{\good} \np{ \util^{\good} - \util^{\bad} } + \util^{\bad} \) is
  increasing as a consequence of
\( \util^{\good} > \util^{\bad} \) by~\eqref{eq:payoffs}}
&=
\int_{\SIMPLEX^1} \bc{ \proba^{\bad}\util^{\bad} + 
\proba^{\good}\util^{\good} }
  \InformationState\bp{d\np{\proba^{\bad},\proba^{\good}}}
\tag{by the definition~\eqref{eq:mean_InformationState} of~$\ic{\InformationState}^{\good}$}
\\
&=
\int_{\SIMPLEX^1} \coutint\np{\state,\Try} \InformationState\np{d\state}
\tag{by~\eqref{eq:one-stage_payoff}.}
\end{align*}
We can prove in the same way that 
\( \VALUE \geq  \VALUE \circ \theta^{\bad} \), 
so that we have obtained
\begin{equation}
\VALUE \circ \theta^{\good} \geq \VALUE \geq \VALUE \circ \theta^{\bad} 
\eqfinp
\label{eq:stay-with-a-winner_not_with_a_looser}
\end{equation}
\medskip

The rest of the proof follows from Proposition~\ref{pr:Optimal_strategy_under_uncertainty}.
In particular, once the optimal~DM selects the ``avoid'' option, 
the~DM will never more experiment.  
Indeed, the optimal rule of 
Proposition~\ref{pr:Optimal_strategy_under_uncertainty} states that, 
once the optimal~DM selects the ``avoid'' option, 
the~DM does not observe the random outcomes, hence 
the~DM no longer updates the posterior~$\InformationState_t$ 
because of the dynamics~\eqref{eq:dynamics} so that the~DM keeps avoiding. 

%
\medskip

This ends the proof. 
\end{proof}

\subsection{Monotonicity property \wrt\ the probability~$\proba^{\bad}$}

We consider the set of functions
\begin{equation}
\mathcal{Z} = \bset{ \varphi : \Delta(\SIMPLEX^1) \to \RR_+ }%
  { \varphi \mtext{ measurable and } \varphi \circ \theta^{\good} \geq \varphi \circ \theta^{\bad} }
\eqfinv
\label{eq:Z}
\end{equation}
where the shift mappings \( \theta^{\bad}, \theta^{\good}  : \Delta(\SIMPLEX^1)
\to \Delta(\SIMPLEX^1) \) have been defined in~\eqref{eq:mappings_theta}.

\begin{proposition}
For any sequence \( \sequence{\varphi_s}{s=0,\ldots,t} \) of functions in~\(
\mathcal{Z} \),
the function \( [0,1] \ni \proba^{\bad} \mapsto 
\EE_{\Bernoulli{\proba^{\bad}}{1-\proba^{\bad}}} 
\bc{ \prod_{s=0}^t \varphi_s\np{\InformationState_s} } \) 
is nonincreasing, 
where, for any \( \np{\proba^{\bad},\proba^{\good}} \in \SIMPLEX^1 \), 
the probability distribution~\(
\Bernoulli{\proba^{\bad}}{\proba^{\good}} \)
on the sample space~$\HISTORY$ in~\eqref{eq:universe}
is given in~\eqref{eq:Bernoulli}, and where the sequence 
\( \sequence{\InformationState_s}{s=0,\ldots,t} \) of posteriors
is given by 
\begin{equation}
\InformationState_0 = \prior_0 \mtext{ and } 
\InformationState_{s+1} = 
\dynamics\np{\InformationState_s,\va{\Uncertain}_{s+1}}=
\begin{cases}
  \theta^{\bad}\InformationState_s & \textrm{if } \va{\Uncertain}_{s+1} = \bad 
\eqfinv
\\
  \theta^{\good}\InformationState_s & \textrm{if } \va{\Uncertain}_{s+1} = \good 
\eqfinp
\end{cases}
\label{eq:dynamics_full}
\end{equation}
\label{pr:Monotonicity}
\end{proposition}

\begin{proof}
Let \( \np{\bar\proba^{\bad},\bar\proba^{\good}} \) and
\( \np{\barbar\proba^{\bad},\barbar\proba^{\good}} \) in~$\SIMPLEX^1$
be such that \( \bar\proba^{\bad} \leq \barbar\proba^{\bad} \)
(or, equivalently, that \( \bar\proba^{\good} \geq \barbar\proba^{\good} \)).
We will show, by induction on~\( t \in \NN \), that, for any sequence 
\( \sequence{\varphi_s}{s=0,\ldots,t} \) of functions in~\( \mathcal{Z} \),
as in~\eqref{eq:Z}, we have the inequality
\begin{equation}
\EE_{\Bernoulli{\bar\proba^{\bad}}{\bar\proba^{\good}}} 
\bc{ \prod_{s=0}^t \varphi_s\np{\InformationState_s} } 
\geq
\EE_{\Bernoulli{\barbar\proba^{\bad}}{\barbar\proba^{\good}}} 
\bc{ \prod_{s=0}^t \varphi_s\np{\InformationState_s} }
\eqfinp 
  \label{eq:induction}
\end{equation}
Before that, we need one notation and two preliminary results.
For any function \( \varphi : \Delta(\SIMPLEX^1) \to \RR \), we put
\begin{equation*}
  \bar{P}\varphi 
= \bar\proba^{\bad} \np{\varphi\circ\theta^{\bad}} + 
\bar\proba^{\good} \np{\varphi\circ\theta^{\good}}
\eqsepv
  \barbar{P}\varphi 
= \barbar\proba^{\bad} \np{\varphi\circ\theta^{\bad}} + 
\barbar\proba^{\good} \np{\varphi\circ\theta^{\good}}
\eqfinp
\end{equation*}
On the one hand, from the equalities 
\[
\bar{P}\varphi-\barbar{P}\varphi =
\np{\bar\proba^{\bad}-\barbar\proba^{\bad}}\np{\varphi\circ\theta^{\bad}}
+ \np{\bar\proba^{\good}-\barbar\proba^{\good}} \np{\varphi\circ\theta^{\good}}
=
\np{\bar\proba^{\bad}-\barbar\proba^{\bad}}
\np{\varphi\circ\theta^{\bad}-\varphi\circ\theta^{\good}}
\eqfinv
\]
we readily get that, as  \( \bar\proba^{\bad} -\barbar\proba^{\bad} \leq 0 \),
\begin{equation}
  \varphi \in \mathcal{Z} \implies
\np{\varphi\circ\theta^{\bad}-\varphi\circ\theta^{\good}} \leq 0
 \implies \bar{P}\varphi \geq \barbar{P}\varphi 
\eqfinp
  \label{eq:induction_t=1}
\end{equation}
On the other hand, we have that 
\begin{equation}
  \varphi \in \mathcal{Z} \implies
\bar{P}\varphi \in \mathcal{Z} \mtext{ and }
\barbar{P}\varphi \in \mathcal{Z} 
\eqfinv
  \label{eq:P_preserves_mathcalZ}
\end{equation}
because, if \( \varphi \in \mathcal{Z} \), one has
\begin{align*}
  \np{\bar{P}\varphi}\circ\theta^{\bad} 
&=
\bar\proba^{\bad} \np{\varphi\circ\theta^{\bad}\circ\theta^{\bad}} + 
\bar\proba^{\good} \np{\varphi\circ\theta^{\good}\circ\theta^{\bad}}
\tag{by definition of~$\np{\bar{P}\varphi}\circ\theta^{\bad}$}
\\
&=
\bar\proba^{\bad} \np{\varphi\circ\theta^{\bad}\circ\theta^{\bad}} + 
\bar\proba^{\good} \np{\varphi\circ\theta^{\bad}\circ\theta^{\good}}
\intertext{because \( \theta^{\bad}\circ\theta^{\good} = \theta^{\good}\circ\theta^{\bad}
\) as easily seen from the definitions~\eqref{eq:mappings_theta}}
& \leq
\bar\proba^{\bad} \np{\varphi\circ\theta^{\good}\circ\theta^{\bad}}+ 
\bar\proba^{\good} \np{\varphi\circ\theta^{\good}\circ\theta^{\good}}
\tag{as \( \varphi \in \mathcal{Z} \) and by definition~\eqref{eq:Z} of~$\mathcal{Z}$}
\\
&=
 \np{\bar{P}\varphi}\circ\theta^{\good}
\tag{by definition of~$\np{\bar{P}\varphi}\circ\theta^{\good}$.}
\end{align*}
The same inequality holds true for \( \barbar{P}\varphi \).
\medskip

Now, we can prove the inequality~\eqref{eq:induction} by induction.
For $t=0$, the inequality~\eqref{eq:induction} is true as it is the trivial equality 
\( \varphi_0\np{\prior_0}=\varphi_0\np{\prior_0} \).
Let us suppose that the inequality~\eqref{eq:induction} holds true, 
for any sequence \( \sequence{\varphi_s}{s=0,\ldots,t} \) of functions in~\(
\mathcal{Z} \).
We consider a sequence \( \sequence{\varphi_s}{s=0,\ldots,t,t+1} \) of functions in~\(
\mathcal{Z} \), and we calculate 
\begin{align*}
  \EE_{\Bernoulli{\bar\proba^{\bad}}{\bar\proba^{\good}}} 
\bc{ \prod_{s=0}^{t+1} \varphi_s\np{\InformationState_s} } 
&=
  \EE_{\Bernoulli{\bar\proba^{\bad}}{\bar\proba^{\good}}} 
\Bc{ \prod_{s=0}^{t} \varphi_s\np{\InformationState_s} 
\EE_{\Bernoulli{\bar\proba^{\bad}}{\bar\proba^{\good}}} 
\bc{ \varphi_{t+1}\np{\InformationState_{t+1}} \mid 
\InformationState_s \eqsepv s=0, \ldots, t } }
\\
&=
  \EE_{\Bernoulli{\bar\proba^{\bad}}{\bar\proba^{\good}}} 
\Bc{ \prod_{s=0}^{t} \varphi_s\np{\InformationState_s} 
\np{ \bar{P}\varphi_{t+1} } \np{\InformationState_{t}} } 
\intertext{by~\eqref{eq:dynamics_full} in which the sequence \( \sequence{\Uncertain_s}{s=0,\ldots,t+1} \) is
  i.i.d. under the probability distribution~\(
\Bernoulli{\proba^{\bad}}{\proba^{\good}} \) in~\eqref{eq:universe} with 
probability $\proba^{\bad}$ (resp. $\proba^{\good}$) to take the value~$\bad$
  (resp.~$\good$)}
& \geq
  \EE_{\Bernoulli{\bar\proba^{\bad}}{\bar\proba^{\good}}} 
\Bc{ \prod_{s=0}^{t} \varphi_s\np{\InformationState_s} 
\np{ \barbar{P}\varphi_{t+1} } \np{\InformationState_{t}} } 
\tag{by~\eqref{eq:induction_t=1}}
\\
& =
  \EE_{\Bernoulli{\bar\proba^{\bad}}{\bar\proba^{\good}}} 
\Bc{ \prod_{s=0}^{t-1} \varphi_s\np{\InformationState_s} 
\times \bp{ \varphi_{t}
\np{ \barbar{P}\varphi_{t+1} } } \np{\InformationState_{t}} } 
\\
& \geq
 \EE_{\Bernoulli{\barbar\proba^{\bad}}{\barbar\proba^{\good}}} 
\Bc{ \prod_{s=0}^{t-1} \varphi_s\np{\InformationState_s} 
\times \bp{ \varphi_{t}
\np{ \barbar{P}\varphi_{t+1} } } \np{\InformationState_{t}} } 
\intertext{by the induction inequality~\eqref{eq:induction} because 
\( \varphi_s\in\mathcal{Z} \) for $s=0,\ldots,t$ by assumption, 
that \(  \barbar{P}\varphi_{t+1}
\in\mathcal{Z} \) by~\eqref{eq:P_preserves_mathcalZ} as 
\( \varphi_{t+1}\in\mathcal{Z} \)  by assumption, and that 
a product of nonnegative functions in~$\mathcal{Z}$ is also in~$\mathcal{Z}$}
&=
  \EE_{\Bernoulli{\bar\proba^{\bad}}{\bar\proba^{\good}}} 
\bc{ \prod_{s=0}^{t+1} \varphi_s\np{\InformationState_s} } 
\tag{by going backward in the same way.}
\end{align*}
This ends the proof.
\end{proof}

\subsection{Proposition~\ref{pr:Optimal_strategy_under_uncertainty_BIAS}}
\label{Optimal_strategy_under_uncertainty_BIAS}

The following Proposition~\ref{pr:Optimal_strategy_under_uncertainty_BIAS} 
details what are the estimates~\( \ic{\InformationState_t}^{\bad} \) of the 
objective probability value~$\bar\proba^{\bad}$ that the optimal~DM is forming 
during the course of learning, and how she/he assesses the environment.
It also establishes how the probability of relevant events
monotonically depends upon objective probabilities.
To our knowledge, these results are new.


\begin{proposition}
Let \( (\va{\Uncertain}_1,\va{\Uncertain}_2,\ldots) \) be a 
sequence of independent Bernoulli trials 
governed by the objective probability distribution~\( \PP^{\delta_{\np{\bar\proba^{\bad},\bar\proba^{\good}}}}=\Bernoulli{\bar\proba^{\bad}}{\bar\proba^{\good}} \),
as in~\eqref{eq:Bernoulli}, 
where \( \np{\bar\proba^{\bad},\bar\proba^{\good}} \in \SIMPLEX^1 \).
Suppose that the~DM adopts the corresponding strategy~$\strategy\opt$
of Proposition~\ref{pr:Optimal_strategy_under_uncertainty},
based on the observations~\( (\va{\Observation}_1,\va{\Observation}_2,\ldots) \) 
inductively given by~\eqref{eq:strategy}
(for $\strategy=\strategy\opt$) and on the sequence of posteriors~\(
\InformationState_t \) given by the dynamics~\eqref{eq:dynamics}.

Suppose also that the~DM holds the prior 
beta distribution~$\prior_0=\beta(n^{\bad}_{0},n^{\good}_{0})$,
where \( n^{\bad}_{0} >0 \) and \( n^{\good}_{0} >0\) are two positive scalars.
Then, if we define the numbers~$N_t^{\bad}$ and $N_t^{\good}$ 
of bad and good outcomes up to stage~$t$ by
\begin{equation}
N_0^{\bad}=N_0^{\good}=0 \eqsepv 
N_t^{\bad} = \sum_{s=1}^t \1_{\{\va{\Observation}_s=\bad\}} \eqsepv
N_t^{\good} = \sum_{s=1}^t \1_{\{\va{\Observation}_s=\good\}} \eqsepv t=1,2,\dots
\label{eq:number}
\end{equation}
the posterior~$\InformationState_t$ 
in Proposition~\ref{pr:Optimal_strategy_under_uncertainty} 
is the beta distribution
\begin{equation}
\InformationState_t =
\beta(n^{\bad}_{0}+N_t^{\bad},n^{\good}_{0}+N_t^{\good})  
\eqfinv 
\label{eq:beta_t}
\end{equation}
on the simplex~$\SIMPLEX^1$, whose conditional
expectation~\eqref{eq:mean_InformationState} 
is given by the statistics 
\begin{equation}
\ic{\InformationState_0}^{\bad} = 
\frac{n^{\bad}_{0} }{n^{\bad}_{0} + n^{\good}_{0} } \mtext{ and }
\ic{\InformationState_t}^{\bad} = 
\frac{n^{\bad}_{0} + N_t^{\bad}}{n^{\bad}_{0} + n^{\good}_{0} + t}
\eqsepv t=1, 2\ldots
\eqfinp
\label{eq:information_state_statistics}
\end{equation}  
Moreover, here are the assessments of the 
objective probability value~$\bar\proba^{\bad}$
and of the environment made by the above optimal~DM.
\begin{enumerate}[a)]
 \item 
\label{it:tau=+infty}
\emph{Infinite learning $\tau=+\infty$.}
Infinite learning can only happen when \( \prior_0 \in \INFORMATIONSTATE_{\Try} \).
When $\tau=+\infty$, the optimal~DM experiments forever
and, the statistics~\( \ic{\InformationState_t}^{\bad} \) 
in~\eqref{eq:information_state_statistics} asymptotically 
reaches the objective probability value~$\bar\proba^{\bad}$, almost surely
under the objective probability distribution~$\PP^{\delta_{\np{\bar\proba^{\bad},\bar\proba^{\good}}}}$, that is, 
\begin{subequations}
\begin{equation}
 \lim_{t\to +\infty} \ic{\InformationState_t}^{\bad} = \bar\proba^{\bad} \eqsepv 
\end{equation}
or, in more precise terms, 
\begin{equation}
  \PP^{\delta_{\np{\bar\proba^{\bad},\bar\proba^{\good}}}} \Ba{
\lim_{t\to +\infty} \ic{\InformationState_t}^{\bad} = \bar\proba^{\bad} \eqsepv 
\tau=+\infty } = \PP^{\delta_{\np{\bar\proba^{\bad},\bar\proba^{\good}}}} \ba{\tau=+\infty } 
\eqfinp
\end{equation}
\end{subequations}
Then, the optimal~DM asymptotically makes an accurate assessment of the
objective best option as \( \lim_{t\to +\infty} \bp{ \ic{\InformationState_t}^{\bad}\util^{\bad} +
  \ic{\InformationState_t}^{\good}\util^{\good} } =
\bar\proba^{\bad}\util^{\bad} + \bar\proba^{\good}\util^{\good} \).

Infinite learning happens with probability 
\( \PP^{\delta_{\np{\bar\proba^{\bad},\bar\proba^{\good}}}}\bp{ \InformationState_t \in \INFORMATIONSTATE_{\Try}
\eqsepv \forall t=0,1,2\ldots } \), which goes up to~1 when the objective 
probability~\( \bar\proba^{\bad} \) of the bad outcome~$\bad$ goes down to~0.
As a consequence, an accurate estimation of the objective probability
of a rare bad outcome is likely.

  \item 
\label{it:tau=0}
\emph{No learning $\tau=0$.}
No learning happens if and only if \( \prior_0 \in \INFORMATIONSTATE_{\Avoid} \).
When $\tau=0$, the optimal~DM never experiments and the~DM initial
estimate~$\ic{\InformationState_0}^{\bad}$ of the  
objective probability value~$\bar\proba^{\bad}$ satisfies
\begin{equation}
  \Critical \leq \ic{\InformationState_0}^{\bad} \eqfinv 
\label{eq:bounds}
\end{equation}
where the critical probability~\( \Critical \) 
is defined in~\eqref{eq:critical}.
From the start, the optimal~DM assesses
that the environment is prone to prudence, as
\( \ic{\InformationState_{0}}^{\bad}\util^{\bad} +
\ic{\InformationState_{0}}^{\good}\util^{\good} \leq \util_{\Avoid} \). 
\item 
\label{it:0<tau<+infty}
\emph{Finite learning $1 \leq \tau<+\infty$.}
Finite learning can only happen when
\( \prior_0 \in \INFORMATIONSTATE_{\Try} \).
When $1 \leq \tau<+\infty$, the optimal~DM experiments till
stage~$\tau$ and the~DM stops her/his estimation of the  
objective probability value~$\bar\proba^{\bad}$ at a
value~\(\ic{\InformationState_{\tau}}^{\bad} \)
which satisfies
\begin{subequations}
\begin{equation}
  \Critical \leq \ic{\InformationState_{\tau}}^{\bad}  
  \eqfinv
\label{eq:bounds}
\end{equation}
or, in more precise terms, 
\begin{equation}
  \PP^{\delta_{\np{\bar\proba^{\bad},\bar\proba^{\good}}}} \Ba{
    \Critical \leq \ic{\InformationState_{\tau}}^{\bad}  \eqsepv
    1 \leq \tau<+\infty } =
  \PP^{\delta_{\np{\bar\proba^{\bad},\bar\proba^{\good}}}} \ba{1 \leq \tau<+\infty } 
  \eqfinp
\end{equation}
\end{subequations}
When the optimal~DM stops experimenting, she/he assesses
that the environment is prone to prudence, as
 \( \ic{\InformationState_{\tau}}^{\bad}\util^{\bad} +
\ic{\InformationState_{\tau}}^{\good}\util^{\good} \leq \util_{\Avoid} \).

Finite learning happens with probability 
\( \PP^{\delta_{\np{\bar\proba^{\bad},\bar\proba^{\good}}}}\bp{ \exists t=1,2,\ldots \eqsepv \InformationState_t \in 
  \INFORMATIONSTATE_{\Avoid} } \),
which goes down to~0 when the objective 
probability~\( \bar\proba^{\bad} \) of the bad outcome~$\bad$ goes down to~0.
As a consequence, 
if the objective probability~\( \bar\proba^{\bad} \) of the \emph{bad} outcome~$\bad$ is 
low enough, in the sense that \( \bar\proba^{\bad} \leq \Critical \), 
when the experiment phase ends at~$\tau<+\infty$, we have 
\begin{equation}
\bar\proba^{\bad} \leq \Critical \leq \ic{\InformationState_{\tau}}^{\bad}  
\eqsepv 
\end{equation}
hence the~DM will \emph{overerestimate}
the objective probability~\( \bar\proba^{\bad} \) of the bad
outcome~$\bad$,
but this with a vanishing probability as~\( \bar\proba^{\bad} \downarrow 0 \).
\end{enumerate}
\label{pr:Optimal_strategy_under_uncertainty_BIAS} 
\end{proposition}


\begin{proof} 
By the dynamics~\eqref{eq:dynamics}, we easily establish that~\eqref{eq:beta_t}
holds true. Equation~\eqref{eq:information_state_statistics} follows from
property of beta distributions~\eqref{eq:beta}.

  \begin{enumerate}[a)]
  \item 
By Proposition~\ref{pr:Optimal_strategy_under_uncertainty},
when $\tau=+\infty$ 
it is optimal to select decision~$\Try$ and experiment forever.
Thus, the observations~\( (\va{\Observation}_1,\va{\Observation}_2,\ldots) \) 
in~\eqref{eq:strategy} coincide with \( (\va{\Uncertain}_1,\va{\Uncertain}_2,\ldots) \),
and we get that 
\( N_t^{\bad} = \sum_{s=1}^t \1_{\{\va{\Uncertain}_s=\bad\}} \) and
\( N_t^{\good} = \sum_{s=1}^t \1_{\{\va{\Uncertain}_s=\good\}} \), for all
$t=1,2,\ldots$, by~\eqref{eq:number}.
As the random variables \( (\va{\Uncertain}_1,\va{\Uncertain}_2,\ldots) \) 
in~\eqref{eq:coordinate_mappings} are i.i.d. under~$\PP^{\delta_{\np{\bar\proba^{\bad},\bar\proba^{\good}}}}$, 
by the Law of large numbers we have that 
\[
\frac{n^{\bad}_{0} + N_t^{\bad}}
{n^{\bad}_{0} + N_t^{\bad} + n^{\good}_{0} + N_t^{\good}}
=
\frac{n^{\bad}_{0} + \sum_{s=1}^t \1_{\{\va{\Uncertain}_s=\bad\}} }
{n^{\bad}_{0} + n^{\good}_{0} + t}
 \to_{t\to +\infty}
\bar\proba^{\bad} \eqsepv 
\PP^{\delta_{\np{\bar\proba^{\bad},\bar\proba^{\good}}}}-\textrm{p.s.}
\]
By~\eqref{eq:information_state_statistics}, 
asymptotically the statistics~\( \ic{\InformationState_t}^{\bad} \) 
reaches the objective probability value~$\bar\proba^{\bad}$ 
almost surely under the
probability~$\PP^{\delta_{\np{\bar\proba^{\bad},\bar\proba^{\good}}}}$.

Now, we show that 
the function \( [0,1] \ni \bar\proba^{\bad} \mapsto
\PP^{\delta_{\np{\bar\proba^{\bad},\bar\proba^{\good}}}}\bp{ \InformationState_t \in \INFORMATIONSTATE_{\Try}
  \eqsepv \forall t=0,1,2\ldots } \) is nonincreasing.
For this purpose, it suffices to prove that, for any stage~$t$, the function
\begin{equation}
[0,1] \ni \bar\proba^{\bad} \mapsto
\PP^{\delta_{\np{\bar\proba^{\bad},\bar\proba^{\good}}}}\bp{ \InformationState_s \in \INFORMATIONSTATE_{\Try}
  \eqsepv \forall s=0,1,2\ldots, t }
  \label{eq:inproof}
\end{equation}
is nonincreasing, and then let \( t \to +\infty \). 
Now, the functions \(
\varphi_s\np{\InformationState}= \1_{ \{
\InformationState \in \INFORMATIONSTATE_{\Try} \} } \) (the same function for all
$s=0,1,2\ldots$) satisfy the assumptions
of Proposition~\ref{pr:Monotonicity} because
\begin{align*}
  \np{ \varphi_s\circ \theta^{\good} }\np{\InformationState}
  &=
    \1_{ \{
\theta^{\good}\InformationState \in \INFORMATIONSTATE_{\Try} \} } 
  \\
  &=    
    \1_{ \{ \np{ \VALUE\circ \theta^{\good} }\np{\InformationState} >
    \frac{\util_{\Avoid}}{1-\discount} \} }
\tag{by~\eqref{eq:INFORMATIONSTATE_Try}}
      \\
  & \geq
    \1_{ \{ \np{ \VALUE\circ \theta^{\bad} }\np{\InformationState} >
  \frac{\util_{\Avoid}}{1-\discount} \} } 
   \tag{as \( \VALUE \circ \theta^{\good} \geq \VALUE \circ \theta^{\bad}
    \) by~\eqref{eq:stay-with-a-winner_not_with_a_looser}}
  \\
   &=
    \1_{ \{
\theta^{\bad}\InformationState \in \INFORMATIONSTATE_{\Try} \} } 
\tag{by~\eqref{eq:INFORMATIONSTATE_Try}}
  \\
 & =
    \np{ \varphi_s\circ \theta^{\bad} }\np{\InformationState}
\eqfinp  
\end{align*}
We conclude, using Proposition~\ref{pr:Monotonicity}, that the
function~\eqref{eq:inproof} is nonincreasing.

Finally, we easily establish that
\( \PP^{\delta_{\np{0,1}}} \bp{ \InformationState_t \in \INFORMATIONSTATE_{\Try}
  \eqsepv \forall t=0,1,2\ldots } = 1 \).
Indeed, under the probability \( \PP^{\delta_{\np{0,1}}} \), we have
\( \va{\Uncertain}_t=\good \) for all stage~$t$ almost-surely, hence
\( \InformationState_{t+1} = \theta^{\good} \InformationState_t \) 
by the dynamics~\eqref{eq:dynamics}.
Therefore, we get that
\begin{align*}
\InformationState_t \in \INFORMATIONSTATE_{\Try}
& \implies
\VALUE\np{\InformationState_t} >
  \frac{\util_{\Avoid}}{1-\discount}
\tag{by~\eqref{eq:INFORMATIONSTATE_Try}}
      \\
& \implies
\VALUE\np{\theta^{\good} \InformationState_t}
\geq \VALUE\np{\InformationState_t} >
    \frac{\util_{\Avoid}}{1-\discount}
  \tag{by~\eqref{eq:stay-with-a-winner_not_with_a_looser}}
  \\
  &  \implies
\VALUE\np{\InformationState_{t+1}} >
\frac{\util_{\Avoid}}{1-\discount}
  \tag{ since \( \InformationState_{t+1} = \theta^{\good} \InformationState_t \) }
      \\
& \implies
\InformationState_{t+1} \in \INFORMATIONSTATE_{\Try}
\tag{by~\eqref{eq:INFORMATIONSTATE_Try}}
  \eqfinp 
\end{align*}

Since \( \prior_0 \in \INFORMATIONSTATE_{\Try} \), we deduce that
\( \PP^{\delta_{\np{0,1}}} \bp{ \InformationState_t \in \INFORMATIONSTATE_{\Try}
  \eqsepv \forall t=0,1,2\ldots } = 1 \).

\item 
See the proof below.
\item 
Let us suppose that \( \tau<+\infty \). We have that 
\begin{align*}
\util_{\Avoid}
&=
\np{1-\discount} 
\VALUE\np{\InformationState_{\tau}} 
\tag{by definition~\eqref{eq:tau} of $\tau$ and since \( \tau<+\infty \)}
\\
& \geq 
\ic{\InformationState_{\tau}}^{\bad} \util^{\bad} + 
\ic{\InformationState_{\tau}}^{\good} \util^{\good} 
\tag{by the inequality~\eqref{eq:inequality_value_function_Try}}
\\
&=- \ic{\InformationState_{\tau}}^{\bad}
\bp{\util^{\good} - \util^{\bad} }  + \util^{\good} 
\tag{since \( \ic{\InformationState_{\tau}}^{\bad} +
  \ic{\InformationState_{\tau}}^{\good} =1 \)
by~\eqref{eq:mean_InformationState}.}
\end{align*}
Rearranging the terms, and using~\eqref{eq:critical}, we obtain that 
\( \ic{\InformationState_{\tau}}^{\bad} \geq \Critical 
= \frac{\util^{\good}-\util_{\Avoid}}{\util^{\good}-\util^{\bad}} \).
\end{enumerate}
The rest of the proof follows using the property that,
if \( \InformationState_0 \in \INFORMATIONSTATE_{\Try} \), then 
\[
\PP^{\delta_{\np{\bar\proba^{\bad},\bar\proba^{\good}}}}\bp{ \exists t=1,2,\ldots \eqsepv \InformationState_t \in 
  \INFORMATIONSTATE_{\Avoid} } = 1 - \PP^{\delta_{\np{\bar\proba^{\bad},\bar\proba^{\good}}}}\bp{ \InformationState_t \in \INFORMATIONSTATE_{\Try}
  \eqsepv \forall t=0,1,2\ldots }
\eqfinp
\]
This ends the proof.
\end{proof}

\bibliographystyle{plain}
\bibliography{DeLara,risk,psychology}

\end{document}